\documentclass[preprint,12pt]{elsarticle}

\usepackage{graphicx}
\usepackage{amssymb}

\usepackage{amsmath,epsfig}
\usepackage{color, colortbl}
\usepackage{pgf}

\usepackage{times}
\usepackage{epsfig}
\usepackage{graphicx}
\usepackage{amsmath}
\usepackage{amssymb}
\usepackage{subcaption}
\usepackage{enumitem}
\usepackage[utf8x]{inputenc}
\usepackage{xfrac}
\usepackage{hypernat}

\makeatletter
\providecommand{\doi}[1]{%
  \begingroup
    \let\bibinfo\@secondoftwo
    \urlstyle{rm}%
    \href{http://dx.doi.org/#1}{%
      doi:\discretionary{}{}{}%
      \nolinkurl{#1}%
    }%
  \endgroup
}
\makeatother

\usepackage{xspace}

\makeatletter
\DeclareRobustCommand\onedot{\futurelet\@let@token\@onedot}
\def\@onedot{\ifx\@let@token.\else.\null\fi\xspace}

\def\eg{\emph{e.g}\onedot} 
\def\ie{\emph{i.e}\onedot} 
 
\def\etc{\emph{etc}\onedot}

\makeatother

\usepackage{booktabs}
\usepackage[pdftex,colorlinks=true,linkcolor=darkblue,citecolor=darkred,urlcolor=blue]{hyperref}

\usepackage{lineno}

\begin{document}

\journal{Signal Processing: Image Communication}

\definecolor{Gray}{gray}{0.9}

\def\oursBaseline{ours: baseline }
\def\oursItertrain{ours:$+T$ }
\def\oursItertrainFeedback{ours:$+T,F$ }
\def\oursItertrainEncoder{ours:$+T,I$ }
\def\oursItertrainEncoderFeedback{ours:$+T,I,F$ }
\def\oursItertrainEncoderFeedbackGF{ours:$+T,I,F,G$ }
\def\oursItertrainEncoderFeedbackGFPT{ours:$+T,I,F,G$ w Synth }
\def\oursItertrainEncoderFeedbackGFPTFT{ours:$+T,I,F,G$ w Synth+FT }

\def\XXX{\textcolor{red}{XXXX}}

\NewDocumentCommand{\rot}{O{45} O{1em} m}{\makebox[#2][l]{\rotatebox{#1}{#3}}}%

\begin{frontmatter}

\title{Getting to 99\% Accuracy in Interactive Segmentation}

\author[tcd]{Marco Forte}
\author[adobe]{Brian Price}
\author[adobe]{Scott Cohen}
\author[adobe]{Ning Xu}
\author[tcd]{François Pitié}

\address[tcd]{Trinity College Dublin, Ireland}
\address[adobe]{Adobe Research, San Jose, USA}

\begin{abstract}
Interactive object cutout tools are the cornerstone of the image editing
workflow. Recent deep-learning based interactive segmentation algorithms have
made significant progress in handling complex images and rough binary selections
can typically be obtained with just a few clicks. Yet, deep learning techniques
tend to plateau once this rough selection has been reached. In this work, we
interpret this plateau as the inability of current algorithms to sufficiently
leverage each user interaction and also as the limitations of current
training/testing datasets.

We propose a novel interactive architecture and a novel training scheme that are
both tailored to better exploit the user workflow.  We also show that
significant improvements can be further gained by introducing a synthetic
training dataset that is specifically designed for complex object
boundaries. Comprehensive experiments support our approach, and our network
achieves state of the art performance.

\end{abstract}


\end{frontmatter}


\section{Introduction}
Interactive image segmentation aims at generating a binary mask that delineates
a foreground object of interest from the background.  Unlike in semantic image
segmentation, user interactions are expected and must be exploited. Typically,
the interaction will come in the form of clicks that the user will place to fix
errors in foreground or background areas. In an ideal interactive session, the
main shape is sketched in a few clicks, then smaller local details are
refined with subsequent edits. Exploring and exploiting this interactive editing
loop is the objective of this paper.

Classic approaches to interactive
segmentation~\cite{Boykov01graphCuts,Kass1988,intelligent_scissors} have found
some use in professional applications like Photoshop. Artists require a tool
which is responsive to their edits and flexible enough to allow for a wide range
of shapes.  The work of Liu et al.~\cite{paintselection} (2009) is particularly
interesting in that regard, as they acknowledge the progressive nature of image
editing and pay particular attention to the artist workflow. For instance, they
make sure that changes to the mask stay localised to the user edit, so as not to
undo progress. These early works have however poor performance on textured regions, and
some images may require many clicks just to get a rough selection.

Recently, deep-learning approaches have excelled at producing rough segmentation
masks with minimal user input. Object masks can now be typically be extracted to
90\% accuracy(mean Intersection over Union/mIoU~\cite{pascal-voc-2012}) in under six
clicks~\cite{XuDeepSelection,MahadevanIterativelySegmentation,ManinisDeepSegmentation,BenardInteractiveWild,Hu2019ASegmentation,LiInteractiveDiversity,majumder2019content,Liew2017RegionalNetworks},
even on highly textured images. A caveat of its successes; deep-learning
approaches are also observed to plateau short of 95\% accuracy, even after 10 or
more clicks and are unable to output very detailed masks, even after 20 clicks.

\begin{figure}[t]
    \centering
    \setlength{\tabcolsep}{0.1em}
    \begin{tabular}{ccc}
    \includegraphics[width=.33\linewidth]{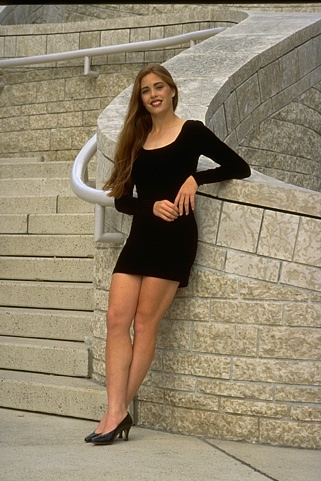} &  
    \includegraphics[width=.33\linewidth]{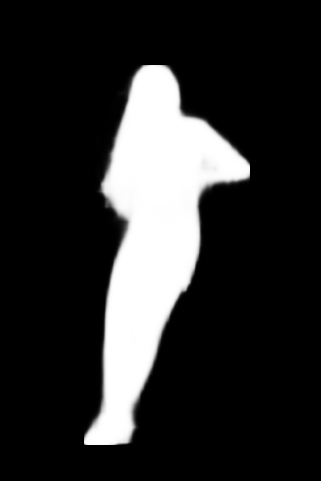} &
    \includegraphics[width=.33\linewidth]{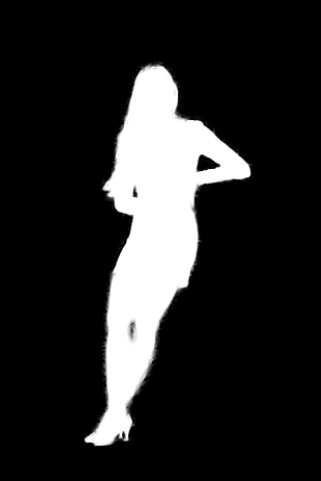} 
    \end{tabular}
    \caption{Comparison between state of the art segmentation at 90\%
      accuracy~\cite{ManinisDeepSegmentation} and our results (right) at 99\%
      accuracy.}\label{fig:90vs99}
\end{figure}

Such low-resolution predictions may be sufficient for casual applications, such
as dataset annotation. For professional high-end applications, such as
Photoshop, this is of little interest as professional artists can manually draw
90\% accurate masks in no time. What they really need is a tool to help them
reach 99-100\% accuracy reliably. See Figure~\ref{fig:90vs99} for a comparison
of results achieving 90\%~\cite{ManinisDeepSegmentation} accuracy and our results which achieve 99\%.

One reason for this plateau is that current architectures manipulate images and
click interactions at low resolution, while fine details are usually only
recovered through post-processing stages (either using graph
cuts~\cite{XuDeepSelection,Liew2017RegionalNetworks} or
CNNs~\cite{jang2019interactive,Hu2019ASegmentation}). This is partly a
consequence of existing benchmarks having set low accuracy targets for the
task~\cite{XuDeepSelection}: in the Berkeley benchmark\cite{BerkeleyDB} the threshold is set to 90\%
accuracy, 90\% for the GrabCut dataset~\cite{grabcutDB} and only 85\% for
PASCAL~\cite{pascal-voc-2012}. These arbitrary targets come from a focus on quick rough selections and imperfections
in the manual delineations of the objects when creating the ground-truth
labelling. Any pixel within 3-5 pixels of the object boundary is typically
ignored in the evaluation. As a result, it was noted in FCN~\cite{FCN} that
working at $8\times$ lower resolution yields sufficient accuracy for the
benchmark tasks of semantic segmentation, and many deep learning architectures are content working at
$8\times$ or $4\times$ lower output resolution than the original
image~\cite{XuDeepSelection,MahadevanIterativelySegmentation,ManinisDeepSegmentation,BenardInteractiveWild,majumder2019content}.

Another reason for the plateau is that current approaches treat all user interactions the
same. During training, a set of foreground and background clicks are typically
placed on the object and background, according to heuristic
strategies~\cite{ManinisDeepSegmentation,Hu2019ASegmentation,LiInteractiveDiversity,majumder2019content,Liew2017RegionalNetworks,MahadevanIterativelySegmentation}. These
clicks are placed together during training, without keeping any state or order information from
one interaction to the next. The networks are thus not specifically trained to
respond to corrective user clicks. Only
in~\cite{BredellIterativeNetworks,MahadevanIterativelySegmentation} was the idea
of training the network one edit at a time was explored, but only after that a
good initial segmentation is given. There is thus a need to better focus on the
way artists approach iterative segmentation and how to respond precisely to
their input.

The primary objective of this paper is to reach higher levels of segmentation
accuracy by working at full resolution and modelling, from the start, the
selection process as a series of interactions, whose purpose shifts over time
from identifying the object globally to refining it locally. To achieve that
goal, we are making a series of contributions covering all aspects of the
problem. These contributions include:
\begin{enumerate}[leftmargin=*]
\item A single architecture which allows for high resolution processing. The core architecture is based on a U-Net style decoder with skip connections to handle fine details. We further incorporate a Guided Filter layer~\cite{Wu2018FastET} to refine the results and produce precise transparency masks.
\item Our architecture is split across two encoding streams: an Image stream and an Interaction stream. This separation of the inputs improves the propagation of the user interactions throughout the network and helps precisely respond to each click.
\item A training strategy where the corrective clicks are sequentially added from the first click, so as to match the artist workflow and thus jointly train for initial rough segmentation and detailed refinements.
\item Demonstrating that a high quality synthetic dataset, specifically designed
  to target fine details, can be used to further improve the quality of the
  segmentation.
\end{enumerate}

The proposed method achieves state of the art performance for any number of
clicks and can reach 99\% accuracy for 62\% of images at 20 clicks. More importantly, it is also demonstrably more usable for artists, as
the network predictions are more predictable and easier to subsequently edit.

The rest of the paper is organised as follows. In Section~\ref{sec:rel_works} we
review prior works related to our approach. In Section~\ref{sec:approach} we
detail our proposed method. In Section~\ref{sec:experiments} we compare our
method to the state of the art algorithms and we perform ablation experiments to
quantitatively and qualitatively assess our contributions. In
Section~\ref{sec:synthetic} we present our new training synthetic dataset and
study how it can impact the quality of the segmentation.

\section{Related Works}\label{sec:rel_works}

In recent years, deep-learning based approaches to interactive segmentation (\eg
~\cite{XuDeepSelection,jang2019interactive,Liew2017RegionalNetworks}) have
superseded prior
works~\cite{intelligent_scissors,paintselection,Boykov01graphCuts,Kass1988}.
While the core architecture of interactive segmentation networks is typically
borrowed from semantic segmentation networks (\eg DeepLab~\cite{DeepLab}), or
general purpose image processing architectures (\eg U-Net~\cite{Unet}), deep
interactive segmentation requires addressing two specific challenges: 1) how to
best encode the sequence of user interactions in the architecture and 2) how to
train the network so as to best match the behaviour of artists. In the rest of
this section, we look at how existing works approach the design and training of
interactive segmentation networks.

\subsection{Segmentation Architectures}

Most previous interactive segmentation networks are either based on semantic
segmentation networks, such as DeepLab~\cite{DeepLab}, or general purpose image
processing networks such as U-Net~\cite{Unet}. Semantic Segmentation networks
typically follow an Encoder-Decoder architecture and use off-the-shelf encoders
pretrained on ImageNet~\cite{imagenet_cvpr09}. For instance, the works of
\citet{XuDeepSelection} and \citet{majumder2019content} are based on the FCN
semantic segmentation network of \citet{FCN}, which itself is based on the
ImageNet trained VGG16 network of \citet{Simonyan14c}. The differences between
notable prior works on Interactive Segmentation are included in
Tables~\ref{tab:priorworks:architectures}, ~\ref{tab:priorworks:interactions}
and~\ref{tab:priorworks:training}.

Note that techniques based on semantic segmentation networks typically operate
at \sfrac{1}{4} or \sfrac{1}{8} resolution, which is a too low resolution for
our aim. Full resolution output can be obtained by using Graph-Cuts (see
iFCN~\cite{XuDeepSelection}, RIS-Net~\cite{Liew2017RegionalNetworks}) or by
appending a dedicated upsampling networks (see FCTSFN~\cite{Hu2019ASegmentation}
and BRS~\cite{jang2019interactive}). Upsampling networks need to be trained
afterwards, in a two-stage process. In RIS-Net~\cite{Liew2017RegionalNetworks},
fullres prediction is obtained by fusing the predictions of two decoder streams:
the first stream makes a global low resolution prediction and the second
stream independently processes patches at click locations to give local hires
predictions. In IIS-LD~\cite{LiInteractiveDiversity}, a single full resolution
network is proposed by using VGG hypercolumns.

\begin{table*}[p]
    \scriptsize
    \begin{tabular}{llm{8em}m{9em}llll}
    \toprule
        Name & Reference & Encoder & Decoder & Scale & Post-Processing \\
        \midrule
        iFCN & Xu et al.~\cite{XuDeepSelection} & VGG16~\cite{Simonyan14c} & FCN~\cite{FCN} & 1:8  & graph-cuts~\cite{Boykov01graphCuts} \\\rowcolor{Gray} 
        ITIS & Mahadevan et al.~\cite{MahadevanIterativelySegmentation} &
        Xception~\cite{chollet2017xception} & DeepLab-v3+~\cite{chen2018encoder} & 1:4  & none \\
        DEXTR & Maninis et al.~\cite{ManinisDeepSegmentation} & ResNet-101~\cite{he2016deep} & Deeplab-v2 & 1:8  & none \\\rowcolor{Gray}
        VOS-wild & Benard et al.~\cite{BenardInteractiveWild} & ResNet-101 & Deeplab-v2~\cite{chen2017deeplab} & 1:8  & CRF~\cite{krahenbuhl2011efficient} \\
        FCTSFN & Hu et al.~\cite{Hu2019ASegmentation} & VGG16 & CNN with
        bilinear upsampling  & 1:1 &  refinement network \\\rowcolor{Gray}
        IIS-LD & Li et al.~\cite{LiInteractiveDiversity} & VGG16 & hypercolumn, CAN~\cite{chen2017fast} & 1:1 &  none \\
        CAMLG & Majumder et al.~\cite{majumder2019content} & VGG16 & FCN & 1:8 &  none \\\rowcolor{Gray}
        RIS-Net & Liew et al.~\cite{Liew2017RegionalNetworks} & VGG16 &
        DeepLab-LargeFOV~\cite{chen2017deeplab}, custom CNN for local predictions around clicks  & 1:1 &  graph-cuts \\
        IITSEN & Bredell et al.~\cite{BredellIterativeNetworks} & U-Net~\cite{Unet} & U-Net & 1:1 & none  \\\rowcolor{Gray}
        BRS & \citet{jang2019interactive} & DenseNet~\cite{huang2017densely} & U-Net & 1:1 & refinement network \\\midrule
        \textbf{Ours} & - & ResNet-50, groupnorm~\cite{groupnorm}, weight standardisation~\cite{weightstandardization} & Pyramid Pooling~\cite{PSPNet},U-Net & 1:1 & guided filter layer~\cite{wu2017fast} \\\bottomrule
    \end{tabular}
    \normalsize
    \caption{Segmentation Architecture in relevant previous works, detailing: the
    type of decoder and encoder used, the scale of the tensor after the decoder and the
    post-processing method used for upsampling the results to full resolution. }
    \label{tab:priorworks:architectures}
\end{table*}

\begin{table*}[p]
    \scriptsize
    \begin{tabular}{lllll}
    \toprule
        Name & Reference & Click Embedding & Click Fusion & Feedback Fusion \\
        \midrule
        iFCN & Xu et al.~\cite{XuDeepSelection} & Distance Map & early & none  \\ \rowcolor{Gray}
        ITIS & Mahadevan et al.~\cite{MahadevanIterativelySegmentation} & Gaussian & early & early \\
        DEXTR & Maninis et al.~\cite{ManinisDeepSegmentation} & Gaussian & early & none \\\rowcolor{Gray}
        VOS-wild & Benard et al.~\cite{BenardInteractiveWild} & Gaussian & early & none \\
        FCTSFN & Hu et al.~\cite{Hu2019ASegmentation} & Distance Map & late & none  \\\rowcolor{Gray}
        IIS-LD & Li et al.~\cite{LiInteractiveDiversity} & Gaussian & late & none \\
        CAMLG & Majumder et al.~\cite{majumder2019content} & Superpixel based embedding~\cite{majumder2019content} & early & optional  \\\rowcolor{Gray}
        RIS-Net & Liew et al.~\cite{Liew2017RegionalNetworks} & Distance Map & early & none  \\
        IITSEN & Bredell et al.~\cite{BredellIterativeNetworks} & Distance Map &  early &  early \\\rowcolor{Gray}
        BRS & \citet{jang2019interactive} & Distance Map & early & none  \\\midrule
        \textbf{Ours} & - & Gaussians at 3 Scales~\cite{LeInteractiveSelection} & both & late  \\\bottomrule
    \end{tabular}
    \normalsize
    \caption{User Interactions Encoding in relevant previous works.}
    \label{tab:priorworks:interactions}
\end{table*}

\begin{table*}[p]
    \scriptsize
    \begin{tabular}[t]{llm{8em}m{16em}m{8em}}
    \toprule
        Name & Reference & Loss & Training Schedule & Training Dataset \\
        \midrule
        iFCN & Xu et al.~\cite{XuDeepSelection} & BCE & all
        clicks at once & PASCAL 2012 \\ \rowcolor{Gray}
        ITIS & Mahadevan et al.~\cite{MahadevanIterativelySegmentation} & BCE & adding one click per epoch & SBD Training + PASCAL 2012 \\
        DEXTR & Maninis et al.~\cite{ManinisDeepSegmentation} &BCE & 4 boundary clicks at once & SBD Full \\ \rowcolor{Gray}
        VOS-wild & Benard et al.~\cite{BenardInteractiveWild} &BCE & all clicks at once~\cite{XuDeepSelection}  & SBD Full \\
        FCTSFN & Hu et al.~\cite{Hu2019ASegmentation} &BCE & all clicks at once~\cite{XuDeepSelection} & PASCAL 2012 \\ \rowcolor{Gray}
        IIS-LD & Li et al.~\cite{LiInteractiveDiversity} & IoU + click location & all clicks at once~\cite{XuDeepSelection} & SBD Training \\
        CAMLG-IIS & Majumder et al.~\cite{majumder2019content} &BCE & all clicks at once~\cite{XuDeepSelection}  & SBD Full \\ \rowcolor{Gray}
        RIS-Net & Liew et al.~\cite{Liew2017RegionalNetworks} &BCE + Click Discounting~\cite{Liew2017RegionalNetworks} & all clicks at once~\cite{XuDeepSelection}  & PASCAL 2012 \\
        IITSEN & Bredell et al.~\cite{BredellIterativeNetworks} &BCE & Separate
                                                                       network
                                                                       makes
                                                                       initial
                                                                       prediction,
                                                                       then adding                                                                        clicks
                                                                       per image.
                                                    &
                                                      Medical~\cite{bredell_medical_dataset} \\ \rowcolor{Gray}
        BRS & \citet{jang2019interactive} & BCE & all clicks at once\cite{jang2019interactive} & SBD Training \\ \midrule
        \textbf{Ours} & - & IoU + click location & adding clicks per image from
        scratch & SBD Training + Synthetic\\\bottomrule
    \end{tabular}
    \normalsize
    \caption{Loss and Training in relevant previous works.}
    \label{tab:priorworks:training}
\end{table*}

\subsection{Embedding User Interactions}

\paragraph{Encoding the Clicks}

The core segmentation network also needs to be adapted to include the user
interactions. In the literature, interactions come in the form of clicks:
positive clicks indicate foreground regions and negative clicks background
regions. Such click inputs are encoded as images either via a distance
transform~\cite{XuDeepSelection,Liew2017RegionalNetworks,BredellIterativeNetworks,Hu2019ASegmentation,majumder2019content}
or by fitting Gaussian masks to each of the
clicks~\cite{ManinisDeepSegmentation,LeInteractiveSelection,BenardInteractiveWild,LiInteractiveDiversity}.

To integrate the click maps into the network, most papers
(iFCN~\cite{XuDeepSelection}, RIS-Net~\cite{Liew2017RegionalNetworks},
ITIS~\cite{MahadevanIterativelySegmentation},
DEXTR~\cite{ManinisDeepSegmentation}) adopt an early fusion scheme by simply
joining the click maps as extra channels to the input image tensor.  Later
fusion schemes are however possible. In IIS-LD~\cite{LiInteractiveDiversity},
the click maps are concatenated with all the upsampled image features
hypercolumn. In FCTSFN (Hu et al.~\cite{Hu2019ASegmentation}), VGG16 is applied
separately to both the input image and the click maps. The lower resolution
features from both inputs are then concatenated before being passed to further
convolutional layers. They show that the late feature fusion of click and image
streams out-performs concatenation of image and clicks at the beginning of a
single stream.

\subsubsection{Mask Predictions Feedback}

In most prior works (including iFCN~\cite{XuDeepSelection},
RISNET~\cite{Liew2017RegionalNetworks}, DEXTR~\cite{ManinisDeepSegmentation},
IIS-LD~\cite{LiInteractiveDiversity}, FCTSFN~\cite{Hu2019ASegmentation}), when a
user adds a new click, the previous prediction is discarded and a new prediction
is make afresh. This means no state is carried on from one click to the next.
We could try, however, to learn from previous clicks and predictions as they
give more context to what the new click is trying to fix. The approach proposed
in ITIS by \citet{MahadevanIterativelySegmentation} and IITSEN by
\citet{BredellIterativeNetworks} is to feed back the mask alongside the image
and the clicks (early fusion) in order to retain some state information. 

One potential issue with this early fusion approach is that it might be hard for
the network to recover from a very poor previous prediction.

\subsection{Training Scheme}
The specificity of training interactive segmentation models is that user clicks
need to be supplied during training. As it is impractical to provide
human-generated clicks, the clicks must be simulated in some way. How best to do
this is an open question.

In their early work on deep interactive segmentation, \citet{XuDeepSelection}
use a set of three strategies for generating clicks from a given segmentation
label. The method goes as follows: a random number of foreground clicks are
placed randomly on the object and a random number of background clicks are
placed near the object. This sampling strategy seeks to mimic typical human
input patterns and it has been adopted by many subsequent
works~\cite{Hu2019ASegmentation,LeInteractiveSelection,LiInteractiveDiversity,majumder2019content,BenardInteractiveWild,Liew2017RegionalNetworks}.

\citet{MahadevanIterativelySegmentation} (ITIS) observe, however, that this way
of predetermining the click placements, independently of the network prediction
errors, has an adverse impact on the final model accuracy. They propose to start
with this scheme at first, but then, at each subsequent epoch, a new click is
added on the centre of the largest incorrect region. This training scheme tries
to increase the correlation between the click locations and prediction
errors. The issue is that the initial click samples are still randomly
grouped. Also, later clicks are only updated one epoch at a time but the network
weights may have changed significantly between each epoch. This means that the
click sequences are not necessarily coherent.

A more direct approach for training interactive image segmentation networks was
taken by Bredell et al.~\cite{BredellIterativeNetworks} for medical image
segmentation. After an initial segmentation is computed by a separate network,
the interactive network is trained by iteratively adding one click at a
time. Each click is placed on the largest incorrect region. The approach,
however, still relies on an initial segmentation, which we believe is
unnecessary.

\section{Proposed Approach}\label{sec:approach}

\subsection{An Interactive Hi-Resolution Network Architecture}\label{sec:architecture}

\begin{figure}[t]
  \centering
  \includegraphics[width=\linewidth]{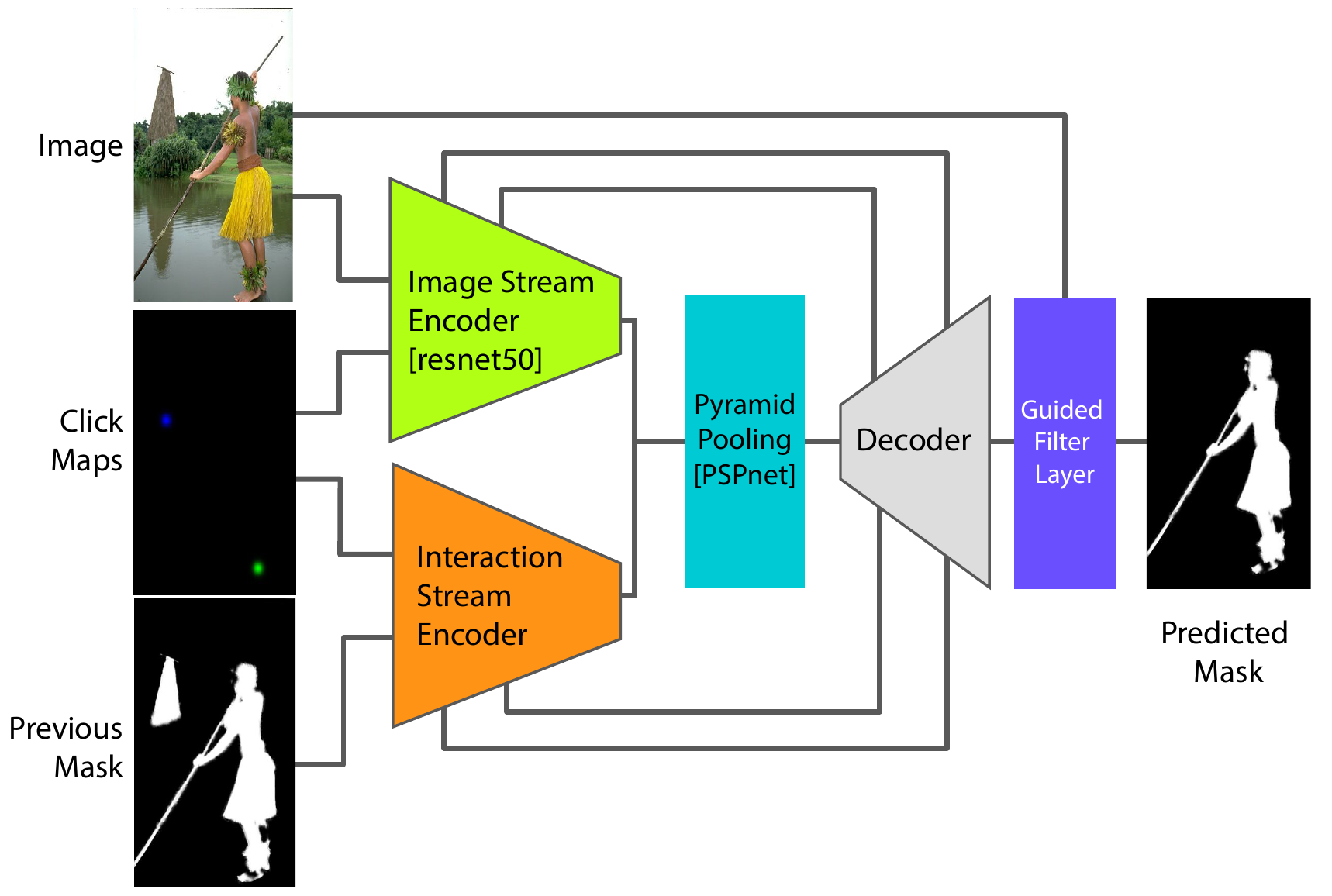}
  \caption{{\bf Proposed Architecture}. The \textit{Image Stream} generates
    image features using ResNet-50, whereas the \textit{Interaction Stream}
    embeds the previous mask estimate and the user clicks. Both streams get fused and pass through the pyramid pooling. The decoder follows a U-Net style architecture. The final layer is a $5\times5$ Deep Guided Filter layer, which allows for fine details extraction. }
  \label{fig:architecture}
  
  \end{figure}

\begin{figure*}[p]

  \begin{tabular}[t]{lm{10em}}
      \includegraphics[width=.7\linewidth]{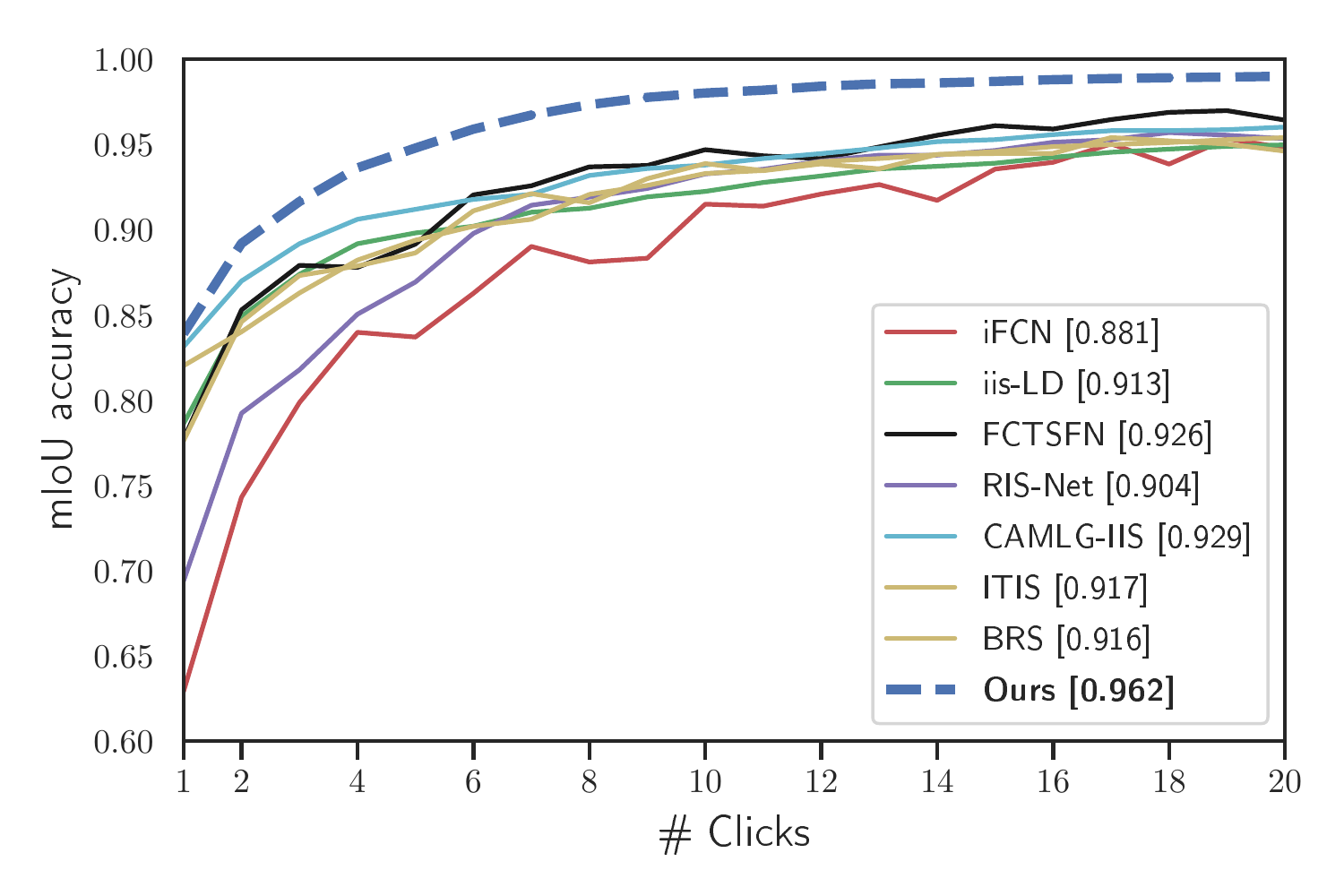} & 
                      {\vspace{-5em} \caption{Comparison of mean IoU scores after $n$ clicks
                          for the {\bf GrabCut} testset~\cite{grabcutDB}.\label{fig:IOUgraphs:GrabCut}}}
                      \\
      \includegraphics[width=.7\linewidth]{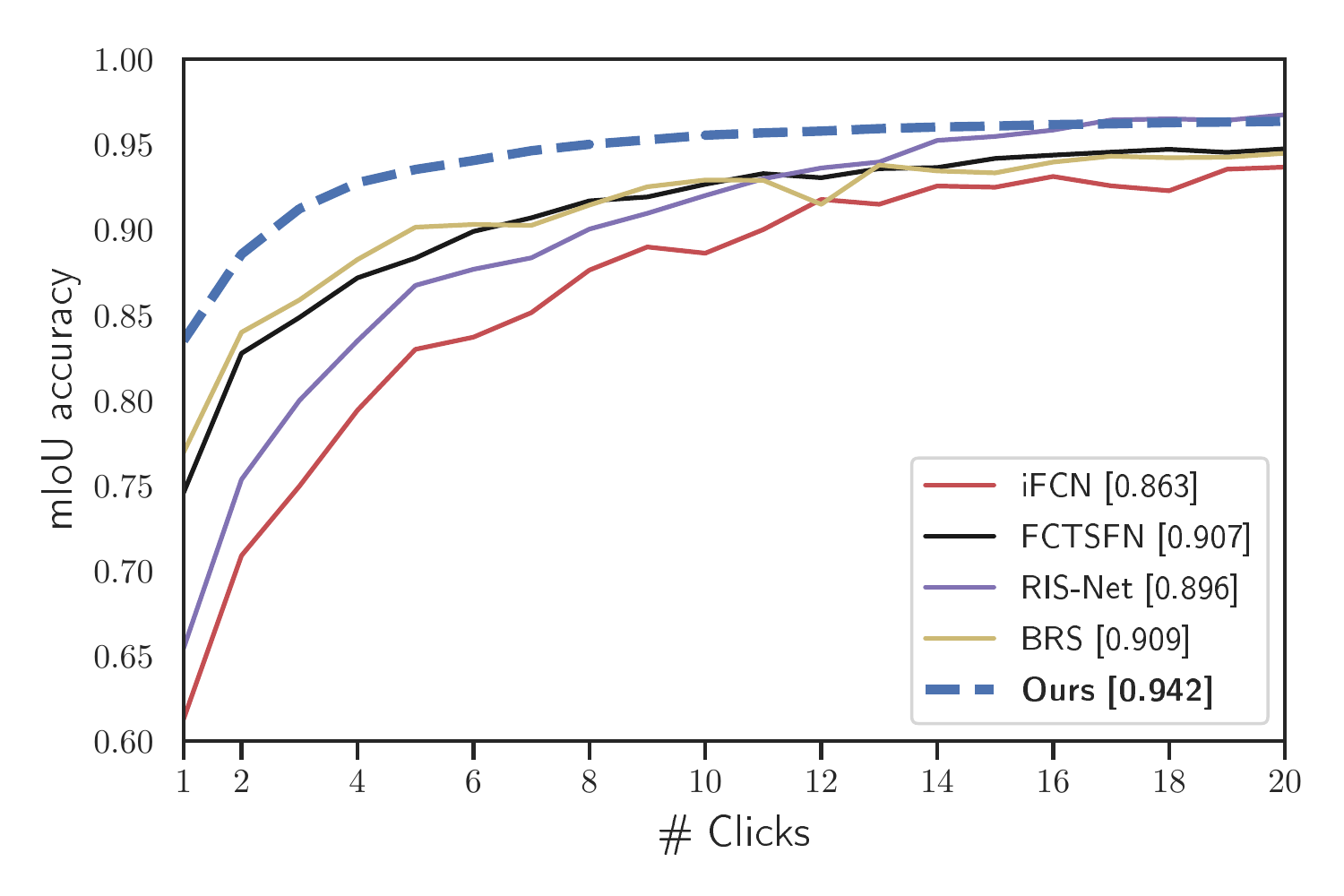} &
                      {\vspace{-5em} \caption{Comparison of mean IoU scores after $n$ clicks
                          for the {\bf Berkeley} testset~\cite{BerkeleyDB}.\label{fig:IOUgraphs:Berkeley}}}
                      \\
      \includegraphics[width=.7\linewidth]{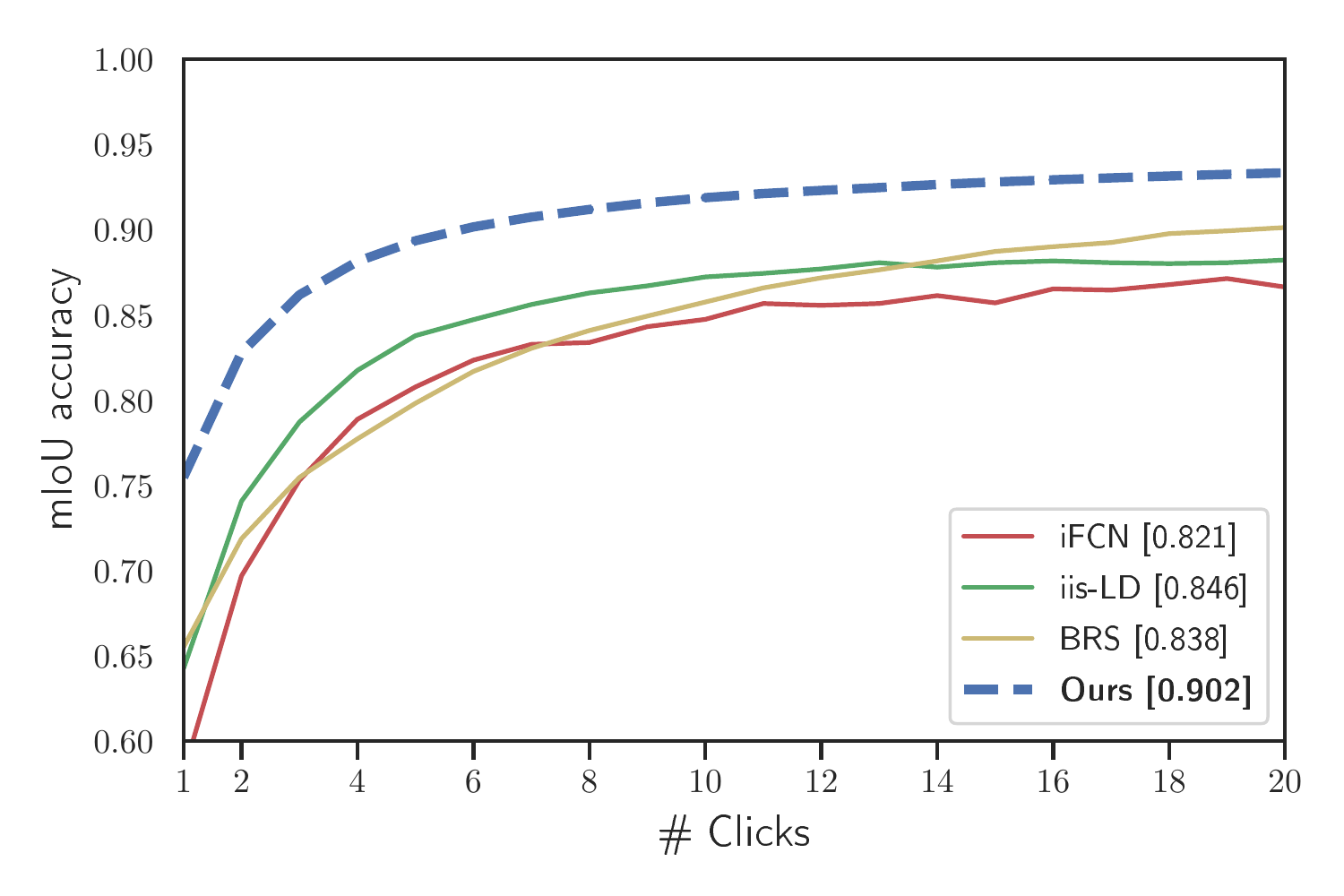} &
                      {\vspace{-5em} \caption{Comparison of mean IoU scores after $n$ clicks
                          for the {\bf SBD} testset~\cite{BharathICCV2011}.\label{fig:IOUgraphs:SBD}}}
  \end{tabular}
\end{figure*}

The proposed architecture is presented in Figure~\ref{fig:architecture}. The
segmentation masks are achieved through the use of an encoder-decoder
network. One key aspect of our model is that we split the input image and user
interactions into two streams: the Image stream and the Interaction stream. The
separation into two streams is designed to allow for both early and late fusion
approaches and thus better respond to user interactions.

Our approach differs in some key ways from prior methods which have used the
previous mask as input to the network~\cite{BredellIterativeNetworks,
  MahadevanIterativelySegmentation} and methods which separate the streams for
clicks and the image~\cite{Hu2019ASegmentation}. Firstly, we adopt a late fusion
for integrating the mask to the image stream. By using a separate branch, we
prevent poor previous mask estimates from affecting the computation of the image
features, while still allowing the mask information to be used. Secondly, we
feed the clicks to both the image and interaction stream, which we observed to
improve accuracy, as opposed to using early or late fusion alone.

\vspace{0.5em}\noindent {\bf The Image Stream} architecture is based on ResNet-50 with
Group Normalisation and Weight Standardisation~\cite{weightstandardization}. The
weights are initialised by pre-training on ImageNet for classification~\cite{weightstandardization}. Two modifications are made
to the network. First, we increase the number of input channels from 3 to 9 to
allow for the extra click input maps. We encode the foreground and background
clicks as in~\cite{LeInteractiveSelection} using Gaussian masks at three
different scales, centred on each click. Second, we remove the striding from
`layer 3' and `layer 4' of ResNet-50 and increase the dilation to 2 and 4
respectively. This increases the output resolution from the Image Stream by 4
times. We experimented with replacing ResNet-50 with ResNet-101 or ResNet-101
pretrained for semantic segmentation, but we did not observe any noticeable
improvement.

Because our training regime requires a batch size of 1, a crucial tweak is
that we use Group Normalisation (GN)~\cite{groupnorm} (32 channels per group)
with Weight Standardisation~\cite{weightstandardization} instead of Batch
Normalisation.

\vspace{0.5em}\noindent \textbf{The Interaction Stream} encodes the previous
mask together with the click maps. It has six convolutional layers (see Table~\ref{tab:ClickMaskEncoder}), each
followed by the LeakyReLu activation function~\cite{maas2013rectifier}. Three of the layers are strided convolutions with stride two, so the output resolution matches that of the Image Stream.

\vspace{0.5em}\noindent \textbf{The Decoder} concatenates features from both streams and
passes them to a Pyramid Pooling layer~\cite{PSPNet}. The pooled features are
fed into a decoder which contains eight convolutional layers interleaved with
three bilinear upsampling layers and skip connections from both streams (see
Table \ref{tab:decoder}). The output of the decoder is at full resolution. 

\vspace{0.5em}\noindent \textbf{Fine Details} are then extracted by appending a
Guided Filter layer~\cite{wu2017fast}.  This further refines the mask details
and allow us to obtain a transparency mask that can capture subpixel
features, such as hair strands, as well as transparency, such as in motion
blur. This means that the predicted mask is now a proper transparency mask that
can be subsequently used for compositing the selected object onto a different
background.

\vspace{0.5em}

The full implementation details of the network architecture can be found on the
project page\footnote{\label{note1} Code will also be published on GitHub upon
  acceptance.}.

\begin{table}[p] \small
\begin{tabular}{llll}
\toprule
Interaction Stream Encoder Layer & X   & C   & S \\ \midrule
Concat: Clicks \& Previous Mask  & 256 & 7   & - \\ 
conv1 & 256 & 64  & 1 \\ 
conv2 & 128 & 128 & 2 \\ 
conv3 & 128 & 256 & 1 \\ 
conv4 & 64  & 256 & 2 \\ 
conv5 & 32  & 256 & 2 \\ 
conv6 & 32  & 256 & 1 \\\bottomrule 
\end{tabular}

\caption{ Network architecture for the Interaction Stream Encoder. \textbf{X} is
  the output spatial resolution, \textbf{C} is output number of channels,
  \textbf{S} is the convolution stride. All convolutional layers are followed by
  LeakyRelu activations \cite{maas2013rectifier}. }
\label{tab:ClickMaskEncoder}

\end{table}

\begin{table}[p] \small
\begin{tabular}{lllll}\toprule
Decoder Layer                                       & X   & C    & GN & Up \\ \midrule
Concat: Features from PPM   \& Interaction Stream `conv6'                       & 32  & 3328  &  & \\ 
conv1                                       & 32  & 256  & \checkmark  &             \\ 
conv2                                       & 32  & 256  & \checkmark  & \checkmark         \\ 
Concat: ResNet `layer 1'   \& Interaction Stream `conv4'              & 64  & 768  &    &  \\ 
conv3                                       & 64  & 256  & \checkmark  & \checkmark       \\ 
Concat: ResNet `conv 3' \& Interaction Stream `conv2'                & 128 & 384  &    & \\
conv4                                       & 128 & 64   & \checkmark  & \checkmark       \\ 
Concat: Image \& Clicks              & 256 & 73   &    &             \\ 
conv5                                       & 256 & 32   &    &             \\ 
conv6                                       & 256 & 16   &    &             \\ 
conv7: (1x1, clip(0,1)) & 256 & 1    &    &             \\\bottomrule
\end{tabular}
\caption{
Network architecture for the decoder. \textbf{X} is the output spatial resolution, \textbf{C} is output number of channels, 
\textbf{GN} indicates if the LeakyReLu was followed by a Group Normalisation
~\cite{groupnorm} layer, \textbf{Up} indicates if the layer was followed by 2x
bilinear upsampling. All convolutional layers except `conv7' are followed by
LeakyReLu activations.}
\label{tab:decoder}

\end{table}

\subsection{Training}\label{sec:training}
\label{subsec:trainingdeets}

\subsubsection*{Training Schedule}

A fundamental deviation from previous works is that we train our network, image
by image, click by click. Whereas in prior works the clicks are first bundled
together, we propose to introduce the clicks sequentially, starting from a
single click, and adopting the same sequential scheme used to evaluate
interactive segmentation algorithms~\cite{Nickisch2010LearningSystem,
  XuDeepSelection, Liew2017RegionalNetworks, LiInteractiveDiversity,
  LeInteractiveSelection, majumder2019content, BenardInteractiveWild,
  Hu2019ASegmentation}.
For each click, we first look at the previously predicted map, and segment the
mislabelled pixels into connected regions. We place the click at the centre of
the largest incorrect region, so as to maximise the Euclidean distance to both
the region boundary and the sides of the image. For the first click, the
prediction map is set to zero. The loss is then computed after the placement of
each click and the weights are updated through back-propagation before the next
click.

By iterating through each click, we are essentially combining the loss for the
entire range of clicks, hence effectively computing the area under the accuracy
per clicks curve (as seen in
Figures~\ref{fig:IOUgraphs:GrabCut},~\ref{fig:IOUgraphs:Berkeley}
and~\ref{fig:IOUgraphs:SBD}). This has the advantage that our training loss
matches the final metric used for evaluation. Also, this means that we are
optimising for the whole range of clicks.
In practice we found that using 4 clicks per image was optimal. The overall
performance did not improve when placing more than 4 clicks but started to
degrade when using 3 or less clicks.

\subsubsection*{Loss}

The loss for a particular image is defined, as in \citet{LiInteractiveDiversity}
(IIS-LD), as the sum of the soft-IoU loss and a click location loss:
\begin{equation}
L(\alpha) = 1 - \frac{\sum_i \alpha_i\hat{\alpha}_i}{\sum_i
  \max(\alpha_i,\hat{\alpha}_i)} +  \sum_{c \in \mathrm{Clicks}} \left(\alpha_c
- \hat{\alpha}_c\right)^2
\end{equation}
where $\alpha_i$ and $\hat{\alpha}_i$ are the mask prediction and ground truth
mask values at pixel $i$.  We found that using
the soft-IoU led to visually and quantitatively better results than when using
the binary cross entropy (BCE) loss. Segmentation results with BCE tend to
show more spurious isolated regions than with the soft-IoU loss.

\subsubsection*{Hyper-Parameters}

The ResNet-50 is initialised by pre-training on
ImageNet~\cite{weightstandardization}. We train our model with the RAdam
optimiser~\cite{liu2019variance} with a learning rate of $10^{-5}$, with a
single image in each batch. The learning rate is reduced by a factor of 0.1 at
epochs 14, 17 and 20, and training completes after 25 epochs. We apply weight
decay of $0.005, 10^{-5}$ to convolutional weights, and the GN parameters
respectively.  Images are cropped at a fixed size of $352\times 352$ pixels, and
we use horizontal flipping, gamma augmentations, and brightness augmentations.

\section{Evaluation}
\label{sec:experiments}

\subsection{Benchmarks}

We evaluate our work across three publicly available segmentation datasets.
{GrabCut}~\cite{grabcutDB} is a 50 image dataset with 50 ground truth masks,
although pixels in a thin band around the object boundaries are not
labelled. {Berkeley}~\cite{BerkeleyDB} is a 96 image dataset with 100 ground
truth masks, the masks are of the highest quality of all three datasets. {SBD}~\cite{BharathICCV2011} validation set contains 2,820 images
with multiple labels per image. The labels span 20 object classes.

\subsection{Comparison to the State of the Art}

\begin{table}[p]
\centering
\begin{tabular}{m{8em}ccc}
\toprule
Method                & \begin{tabular}[c]{@{}c@{}}GrabCut \\ NoC @ 90\%\end{tabular} & \begin{tabular}[c]{@{}c@{}}Berkeley  \\ NoC @ 90\%\end{tabular} & \begin{tabular}[c]{@{}c@{}}SBD\\ NoC @ 85\%\end{tabular} \\ \midrule
IFCN~\cite{XuDeepSelection}                  & 6.04                                                      & 8.65                                                       & 9.18                                                           \\ \rowcolor{Gray}
RIS-Net~\cite{Liew2017RegionalNetworks}               & 5.00                                                      & 6.03                                                       & -                                                            \\
ITIS~\cite{MahadevanIterativelySegmentation}                  & 5.60                                                      & -                                                          & -                                                            \\\rowcolor{Gray}
DEXTR~\cite{ManinisDeepSegmentation}                 & 4.00                                                      & -                                                          & -                                                           \\
VOS-wild~\cite{BenardInteractiveWild}              & 3.8                                                       & -                                                          & -                                                            \\ \rowcolor{Gray}

FCTSFN~\cite{Hu2019ASegmentation}                & 3.76                                                      & 6.49                                                       & -                                                            \\
IIS-LD~\cite{LiInteractiveDiversity}                & 4.79                                                      & -                                                         & 7.41                                                                \\\rowcolor{Gray}
CAMLG-IIS~\cite{majumder2019content}              &\underline{ 3.58  }                                                    & 5.60                                                       & -                                                           \\
BRS~\cite{jang2019interactive}              & 3.6                                                      & \underline{5.08}                                                       & \underline{6.59}                                                          \\
\midrule
\textbf{\textit{Ours}} & \textbf{2.54}  & \textbf{3.53}  & \textbf{3.90}  \\ \bottomrule
\end{tabular}
\caption{The mean number of clicks required to achieve a certain mIoU accuracy on different datasets~\cite{grabcutDB, BerkeleyDB, BharathICCV2011} by various algorithms. The best results are indicated in bold with the second best underlined.}\label{tab:comparison_num_click} 
\end{table}

In this section, we use two metrics to demonstrate the superior performance of
our algorithm over existing work.

\vspace{0.5em}
\noindent\textbf{Quick Selection}. On Table \ref{tab:comparison_num_click} we
report the mean number of clicks that it takes to reach the customarily used
85\%/90\% IoU threshold, known as the Number of Clicks metric(NoC @ $x$\%). Our algorithm reaches 90\% accuracy in 2.54 clicks on
GrabCut and 3.53 clicks on Berkeley. It reaches 85\% in 3.90 clicks on the SBD
test set. These are all significant improvements over the state of the art. 

We primarily use this metric in order to compare with results reported from previous methods. As seen in Figure~\ref{fig:90vs99} the segmentation quality at 90\% is not sufficient for image editing, and as seen in Figures~\ref{fig:IOUgraphs:GrabCut}--\ref{fig:IOUgraphs:SBD} many methods plateau in accuracy around those thresholds. So we believe it is more important to measure accuracy for a wide range of clicks. 

\vspace{0.5em}
\noindent\textbf{Accuracy per Click}. On Figures
\ref{fig:IOUgraphs:GrabCut}--\ref{fig:IOUgraphs:SBD} is reported the average
segmentation accuracy (mean IoU) across the first 20 iterated clicks for state
of the art algorithms on each dataset. The click placement strategy used in the
evaluation is the same used by all other methods. In the legend we also report
the area under curve (AuC) score across the full range of clicks (1-20). From
the graph and the AuC score we see that our algorithm outperforms all others, on
all datasets, for every number of clicks. On the GrabCut graph we see that for the first click our method matches the accuracy of CAMLG-IIS~\cite{majumder2019content} yet their method does not make good use of the corrective clicks compared to ours. 
We believe the AuC metric is much more relevant for future works than the number of clicks, hence we use it for our ablation study. The graphs show there is still large room for improvement in future work for both at early clicks and for reducing the plateau at later clicks, especially on the Berkeley and SBD datasets.

\subsection{Ablation Study}

We study the effectiveness of our contributions for the following scenarios:

\begin{itemize}
\item A baseline model for our method is based on the Image stream, pyramid
  pooling layer and decoder alone and does not include the Interaction Stream
  nor the Guided Filter Layer.  It is trained using the standard bundled click
  strategy from Ning et al.~\cite{XuDeepSelection}.
\item In ($+T$), we change the training approach to our click by click iterative
  training scheme.
\item In ($+T,+I$), we also include the Interaction Stream ($+I$), without the
  previous mask feedback (\ie it is set to zero).
\item In ($+T,+I,+F$), the previous predicted mask is fed to the Interaction Stream.
\item In ($+T,+I,+F,+G$), our full model is completed by appending a Guided Filter Layer. Unless otherwise stated all results in this paper refer to this model.
\end{itemize}  

\vspace{0.5em}
\noindent\textbf{Accuracy per Click.} On Figures
\ref{fig:AblationIOUgraphs:GrabCut}, \ref{fig:AblationIOUgraphs:Berkeley} and
\ref{fig:AblationIOUgraphs:SBD} are reported the mean IoU across the range of 20
clicks on the SBD dataset~\cite{BharathICCV2011}. The area under the curve, with
its confidence interval, is also compiled in Table~\ref{tab:comparison_mIoU}. We
plot the results from the state of the art method BRS~\cite{jang2019interactive}
as a reference.

We can see that our stripped down baseline model performs about on par with the
state of the art, which points to the strength of our ResNet-50 based Encoder
Decoder architecture with Group Normalisation. Then, training the model with our
interactive regime ($+T$) gives a clear boost to accuracy for all clicks. Then
adding the interactive stream ($+T,+I$) gives a consistent improvement at higher
numbers of clicks, where fine edits are made. Similarly, adding the feedback of
the mask ($+T,+I,+F$) improves results for later clicks on all three
datasets. The Guided Filter layer gives improvements on the
Grabcut~\cite{grabcutDB} and Berkeley~\cite{BerkeleyDB} datasets but not on the
SBD~\cite{BharathICCV2011} set. We believe this is because the labels on SBD do
not follow the object boundaries closely (see discussion in
section~\ref{sec:synthetic}), but Guided Filter refines the prediction based on
the image edges.

\begin{table}[p]
\centering
\begin{tabular}{m{8em}ccc}
\toprule
Method                & GrabCut & Berkeley & SBD \\ \midrule
\oursBaseline                & $0.9121 \pm 0.0078$ & $0.9065 \pm 0.0036$ & $0.8416 \pm 0.0008$   \\ \rowcolor{Gray}
\oursItertrain              & $0.9516 \pm 0.0041$ & $0.9306 \pm 0.0022$ & $0.8842 \pm 0.0006$ \\
\oursItertrainEncoder                 & $0.9486 \pm 0.0038$ & $0.9350 \pm 0.0019$ & $0.8941 \pm 0.0005$\\\rowcolor{Gray}
\oursItertrainEncoderFeedback               & $0.9625 \pm 0.0031$ & $0.9396 \pm 0.0018$ & $0.9079 \pm 0.0005$ \\
\oursItertrainEncoderFeedbackGF             & $0.9628 \pm 0.0025$ & $0.9423 \pm 0.0018$ & $0.9026 \pm 0.0005$  \\ 
\end{tabular}
\caption{ Mean IoU for the 1-20 clicks range on the different datasets~\cite{grabcutDB, BerkeleyDB, BharathICCV2011} for each  variant of our algorithm. }\label{tab:comparison_mIoU} 
\end{table}

\begin{figure}[p]
  \begin{tabular}[t]{lm{12em}}
     \resizebox{0.7\linewidth}{!}{\input{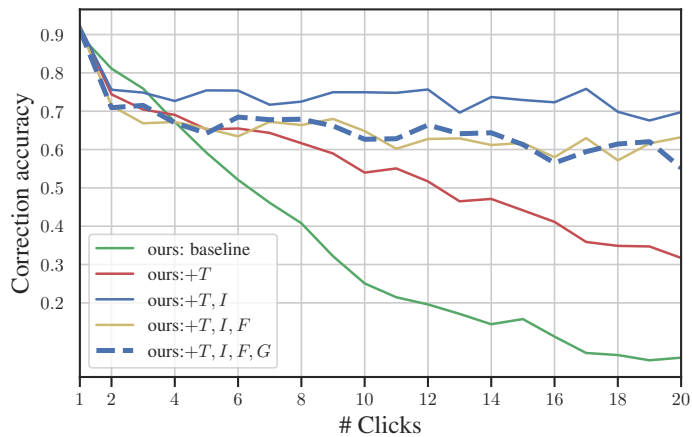}} & 
                      {\vspace{-8em} \caption{Correction accuracy at each click for the Berkeley
      dataset~\cite{BerkeleyDB}. The correction accuracy is defined as the
      proportion of pixels correctly predicted on the region where the click was
      placed (see text).}
                                                                             \label{fig:correctionGraph}}
                                                                             \end{tabular}
\end{figure}

\vspace{0.5em}
\noindent\textbf{Corrective Click Accuracy.} On
Figure~\ref{fig:correctionGraph}, we introduce a novel way to measure
interactive segmentation performance. At each click iteration, we measure the
proportion of correctly predicted pixels within the previous incorrect region,
which is where this new click was placed. In effect, this measures the networks
ability to accurately respond to user inputs, something very important for the
user experience. We compare on Berkeley our baseline model to the one with
interactive training and the full model, which also includes the Guided Filter Layer, and
the interaction stream with mask feedback.

Interestingly, all three methods perform equally for the first three
clicks. However, the baseline model continuously drops in corrective accuracy
after this. The more clicks are added, the less it responds to each click.  Our
interactive training improves this and has much higher corrective accuracy for
all clicks, but it still drops over time.  Our full model maintains a steady
50\% corrective accuracy for each click, indicating it more consistently
leverages each user input.

\begin{figure*}[p]
  \begin{tabular}[t]{lm{10em}}
     \resizebox{0.7\linewidth}{!}{\input{fig/ablation/normal_ablation_grabcut.pgf}} & 
                      {\vspace{-5em} \caption{Comparison of mean IoU scores after $n$ clicks
                          for the {\bf GrabCut} testset~\cite{grabcutDB}.\label{fig:AblationIOUgraphs:GrabCut}}}
                      \\
     \resizebox{0.7\linewidth}{!}{\input{fig/ablation/normal_ablation_berkeley.pgf}} &
                      {\vspace{-5em} \caption{Comparison of mean IoU scores after $n$ clicks
                          for the {\bf Berkeley} testset~\cite{BerkeleyDB}.\label{fig:AblationIOUgraphs:Berkeley}}}
                      \\
     \resizebox{0.7\linewidth}{!}{\input{fig/ablation/normal_ablation_sbd.pgf}} &
                      {\vspace{-5em} \caption{Comparison of mean IoU scores after $n$ clicks
                          for the {\bf SBD} testset~\cite{BharathICCV2011}.\label{fig:AblationIOUgraphs:SBD}}}
  \end{tabular}
\end{figure*}

\vspace{0.5em}
\noindent\textbf{Reaching 99\% accuracy} On Figures
\ref{fig:clicks_distribution:GrabCut}, \ref{fig:clicks_distribution:Berkeley} and
\ref{fig:clicks_distribution:SBD} we plot the proportion of images where the prediction accuracy surpasses 90-99\% accuracy at 1,5,10 and 20 clicks. We see that by adding more clicks we not only raise the lower bar of accuracy but also the number of images reaching 99\%. At 20 clicks, 100\% of images exceed 95\% accuracy, and 62\% of images exceed $99\%$ accuracy on the GrabCut dataset. The Berkeley and SBD datasets are more difficult, yet at 20 clicks our method still exceeds 95\% accuracy for 81\% and 55\% respectively. 

\begin{figure*}[p]
  \begin{tabular}[t]{lm{10em}}
     \resizebox{0.7\linewidth}{!}{\input{fig/clicks_distribution/oursItertrainEncoderFeedbackGF_grabcut.pgf}} & 
                      {\vspace{-10em} \caption{Proportion of images surpassing $x\%$ accuracy after $n$ clicks  for the {\bf GrabCut} testset~\cite{grabcutDB}. \\
                           \textit{62\%} of predictions exceed $99\%$ accuracy after 20 clicks.
                          \label{fig:clicks_distribution:GrabCut}}}
                      \\
     \resizebox{0.7\linewidth}{!}{\input{fig/clicks_distribution/oursItertrainEncoderFeedbackGF_berkeley.pgf}} &
                      {\vspace{-10em} \caption{Proportion of images surpassing $x\%$ accuracy after $n$ clicks  for the {\bf Berkeley} testset~\cite{BerkeleyDB}. \\
                          \textit{81\%} of predictions exceed $95\%$ accuracy after 20 clicks.
                          \label{fig:clicks_distribution:Berkeley}}}
                      \\
     \resizebox{0.7\linewidth}{!}{\input{fig/clicks_distribution/oursItertrainEncoderFeedbackGF_sbd.pgf}} &
                      {\vspace{-10em} \caption{Proportion of images surpassing $x\%$ accuracy after $n$ clicks for the {\bf SBD} testset~\cite{BharathICCV2011}. \\
                           \textit{55\%} of predictions exceed $95\%$ accuracy after 20 clicks.
                          \label{fig:clicks_distribution:SBD}}}
  \end{tabular}
\end{figure*}

\subsection{Qualitative Evaluation}
\label{subsec:qualeval}

\begin{figure*}[p]
 \centering
\setlength{\tabcolsep}{0.1em}
\begin{tabular}{ccccccc}
         & Click 1 & Click 2 & Click 3 & Click 4  & Click 5  & Click 6 \\
BL & \includegraphics[width=.16\linewidth]{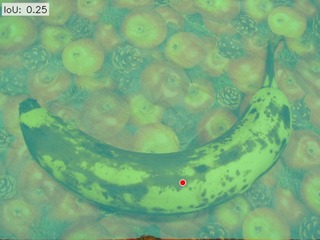}  &  \includegraphics[width=.16\linewidth]{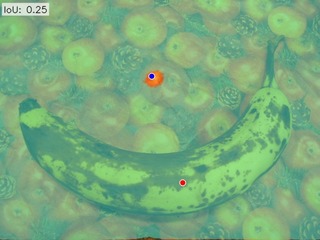} & \includegraphics[width=.16\linewidth]{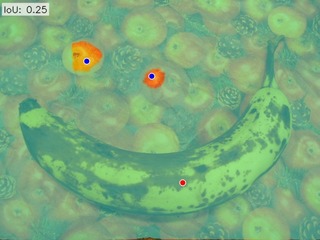}  &  \includegraphics[width=.16\linewidth]{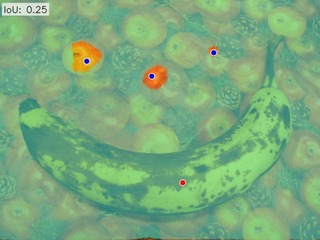}   &  \includegraphics[width=.16\linewidth]{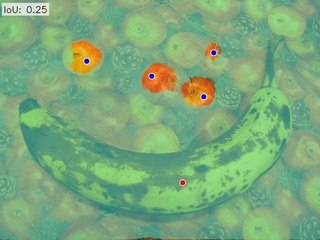}  &  \includegraphics[width=.16\linewidth]{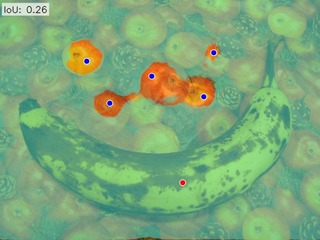} \\ 
Ours & \includegraphics[width=.16\linewidth]{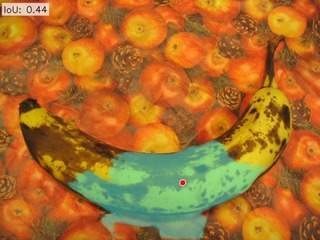}   & \includegraphics[width=.16\linewidth]{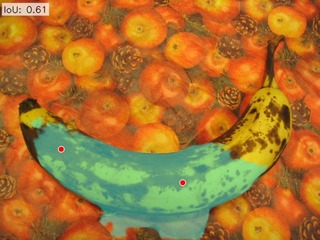}  & \includegraphics[width=.16\linewidth]{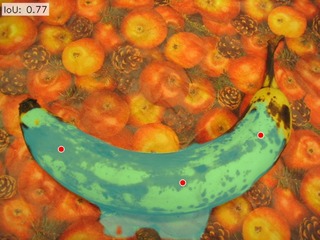}  &   \includegraphics[width=.16\linewidth]{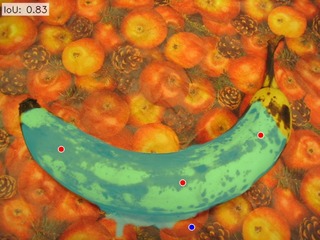} &   \includegraphics[width=.16\linewidth]{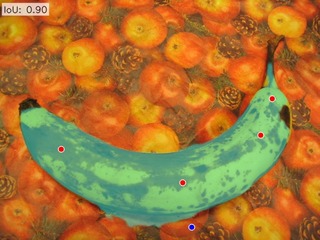} &   \includegraphics[width=.16\linewidth]{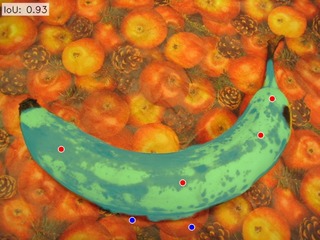}\\

BL & \includegraphics[width=.16\linewidth]{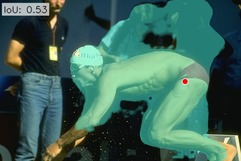}  &  \includegraphics[width=.16\linewidth]{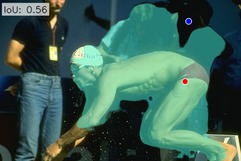} & \includegraphics[width=.16\linewidth]{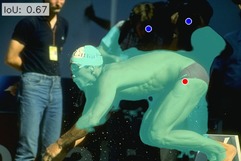}  &  \includegraphics[width=.16\linewidth]{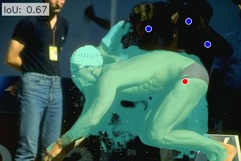}   &  \includegraphics[width=.16\linewidth]{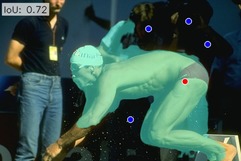}  &  \includegraphics[width=.16\linewidth]{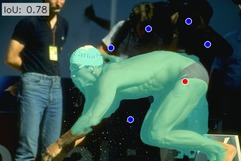} \\ 
Ours & \includegraphics[width=.16\linewidth]{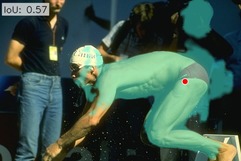}   & \includegraphics[width=.16\linewidth]{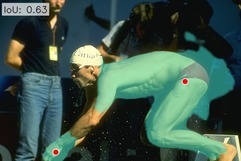}  & \includegraphics[width=.16\linewidth]{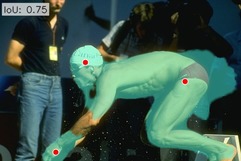}  &   \includegraphics[width=.16\linewidth]{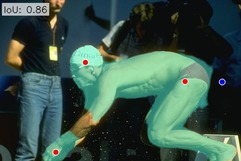} &   \includegraphics[width=.16\linewidth]{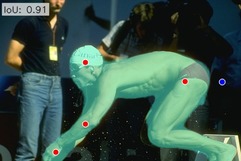} &   \includegraphics[width=.16\linewidth]{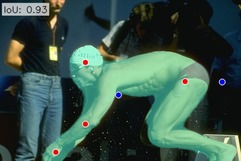}\\

BL & \includegraphics[width=.16\linewidth]{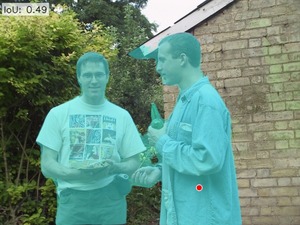}  &  \includegraphics[width=.16\linewidth]{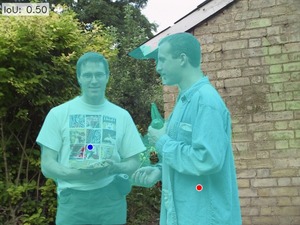} & \includegraphics[width=.16\linewidth]{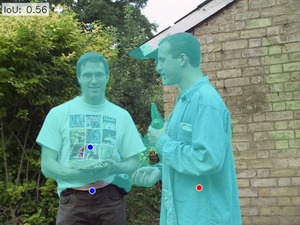}  &  \includegraphics[width=.16\linewidth]{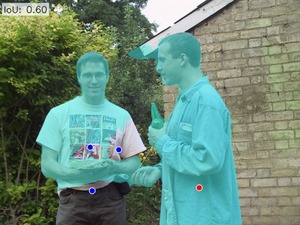}   &  \includegraphics[width=.16\linewidth]{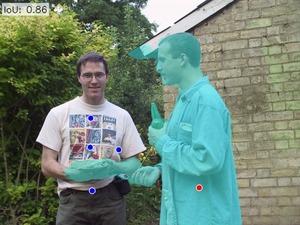}  &  \includegraphics[width=.16\linewidth]{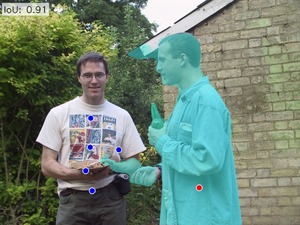} \\ 
Ours & \includegraphics[width=.16\linewidth]{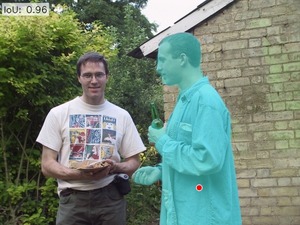}   & \includegraphics[width=.16\linewidth]{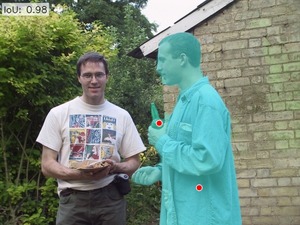}  & \includegraphics[width=.16\linewidth]{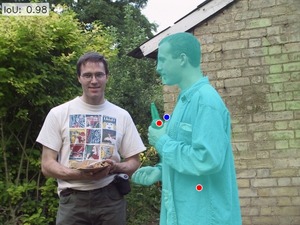}  &   \includegraphics[width=.16\linewidth]{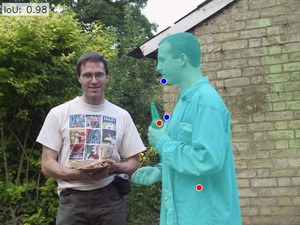} &   \includegraphics[width=.16\linewidth]{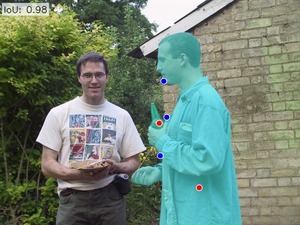} &   \includegraphics[width=.16\linewidth]{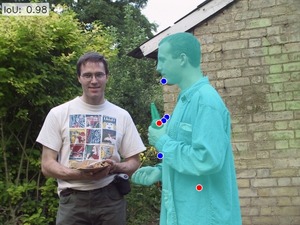}\\
BL & \includegraphics[width=.16\linewidth]{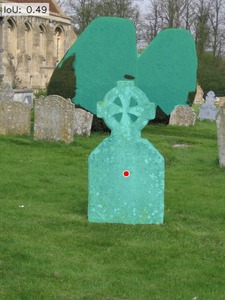}  &  \includegraphics[width=.16\linewidth]{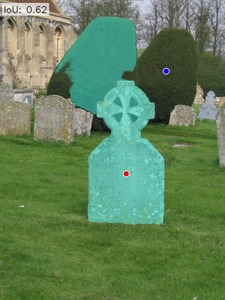} & \includegraphics[width=.16\linewidth]{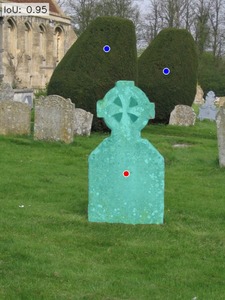}  &  \includegraphics[width=.16\linewidth]{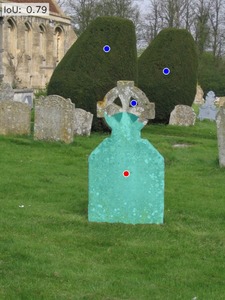} &  \includegraphics[width=.16\linewidth]{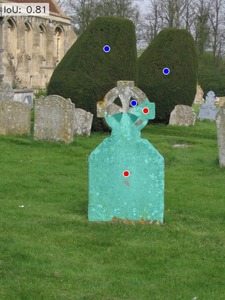} &  \includegraphics[width=.16\linewidth]{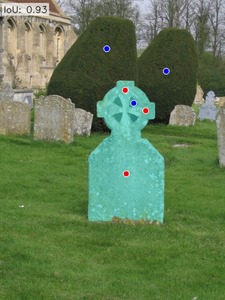}\\ 
Ours & \includegraphics[width=.16\linewidth]{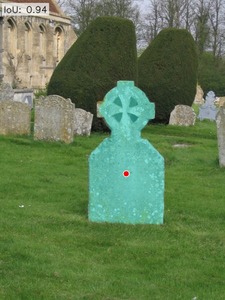}   & \includegraphics[width=.16\linewidth]{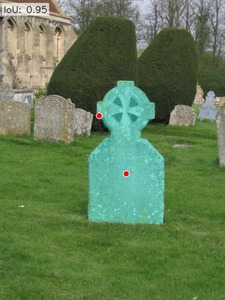}  & \includegraphics[width=.16\linewidth]{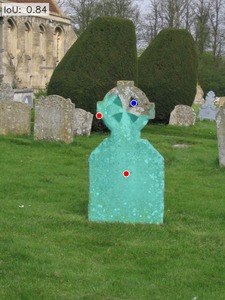}  &   \includegraphics[width=.16\linewidth]{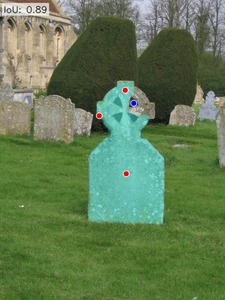} &   \includegraphics[width=.16\linewidth]{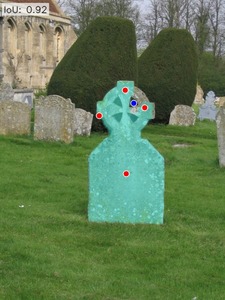} &   \includegraphics[width=.16\linewidth]{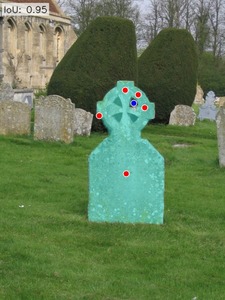}\\
\end{tabular}
\caption{Qualitative comparison between our baseline (BL) and our model across the first 4 clicks. Positive clicks are in red, negative in blue, object masks are highlighted in cyan. Images are from the GrabCut Dataset.}\label{fig:Grabcutcompare_iter_clicks}
\end{figure*}

\begin{figure*}[p]
 \centering
\setlength{\tabcolsep}{0.1em}
\begin{tabular}{ccccccc}
         & Click 1 & Click 2 & Click 3 & Click 4  & Click 5  & Click 6 \\

BL & \includegraphics[width=.16\linewidth]{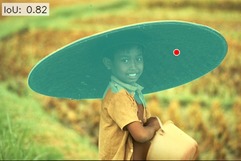}  &  \includegraphics[width=.16\linewidth]{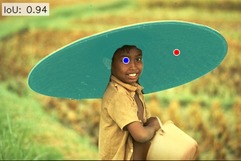} & \includegraphics[width=.16\linewidth]{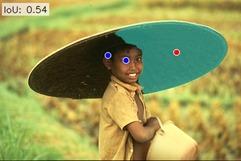}  &  \includegraphics[width=.16\linewidth]{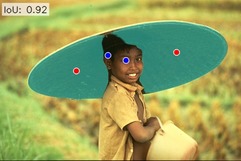}   &  \includegraphics[width=.16\linewidth]{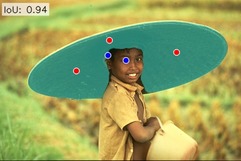}  &  \includegraphics[width=.16\linewidth]{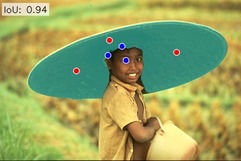} \\ 
Ours & \includegraphics[width=.16\linewidth]{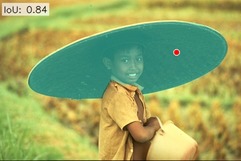}   & \includegraphics[width=.16\linewidth]{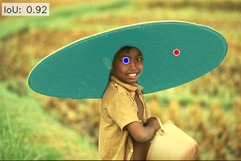}  & \includegraphics[width=.16\linewidth]{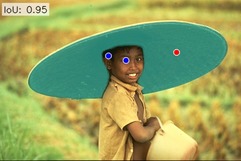}  &   \includegraphics[width=.16\linewidth]{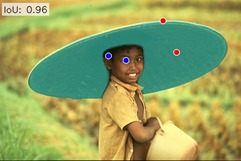} &   \includegraphics[width=.16\linewidth]{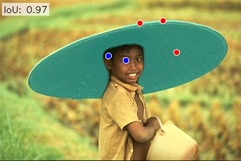} &   \includegraphics[width=.16\linewidth]{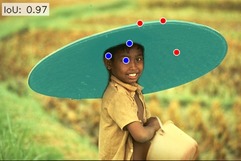}\\
BL & \includegraphics[width=.16\linewidth]{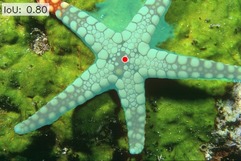}  &  \includegraphics[width=.16\linewidth]{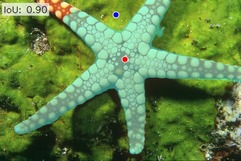} & \includegraphics[width=.16\linewidth]{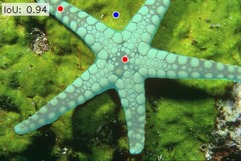}  &  \includegraphics[width=.16\linewidth]{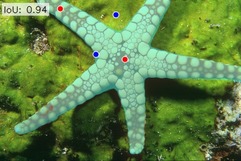} &  \includegraphics[width=.16\linewidth]{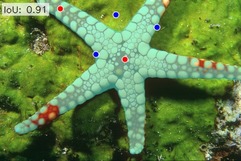} &  \includegraphics[width=.16\linewidth]{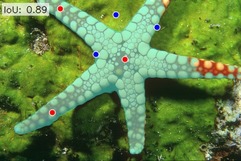}\\ 
Ours & \includegraphics[width=.16\linewidth]{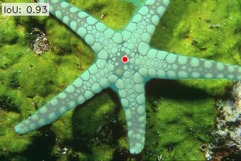}   & \includegraphics[width=.16\linewidth]{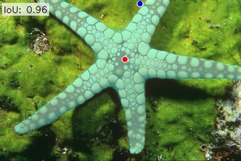}  & \includegraphics[width=.16\linewidth]{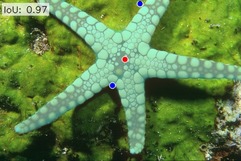}  &   \includegraphics[width=.16\linewidth]{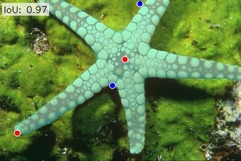} &   \includegraphics[width=.16\linewidth]{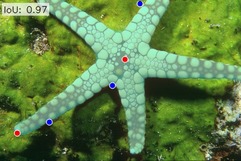} &   \includegraphics[width=.16\linewidth]{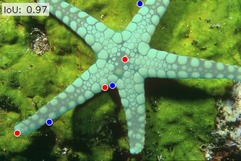}\\
BL & \includegraphics[width=.16\linewidth]{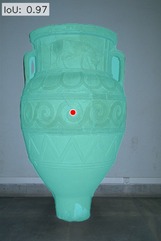}  &  \includegraphics[width=.16\linewidth]{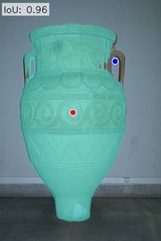} & \includegraphics[width=.16\linewidth]{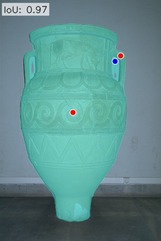}  &  \includegraphics[width=.16\linewidth]{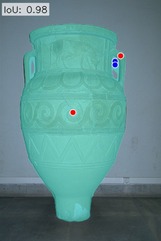}   &  \includegraphics[width=.16\linewidth]{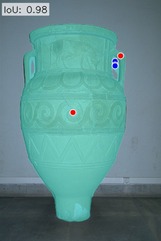}  &  \includegraphics[width=.16\linewidth]{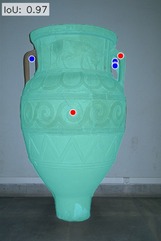} \\ 
Ours & \includegraphics[width=.16\linewidth]{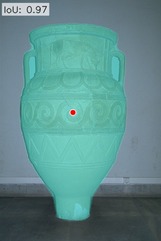}   & \includegraphics[width=.16\linewidth]{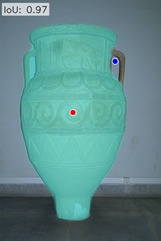}  & \includegraphics[width=.16\linewidth]{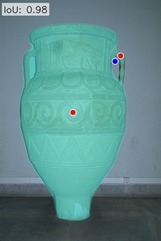}  &   \includegraphics[width=.16\linewidth]{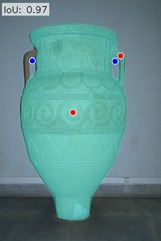} &   \includegraphics[width=.16\linewidth]{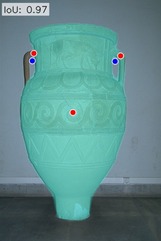} &   \includegraphics[width=.16\linewidth]{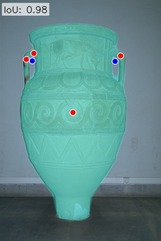}\\

BL & \includegraphics[width=.16\linewidth]{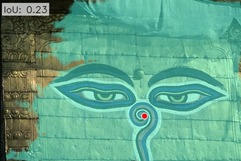}  &  \includegraphics[width=.16\linewidth]{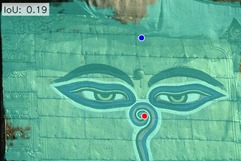} & \includegraphics[width=.16\linewidth]{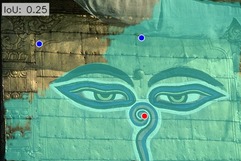}  &  \includegraphics[width=.16\linewidth]{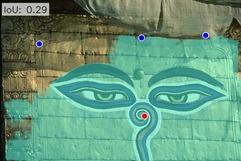} &  \includegraphics[width=.16\linewidth]{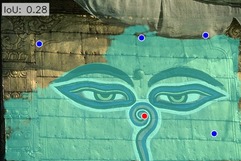} &  \includegraphics[width=.16\linewidth]{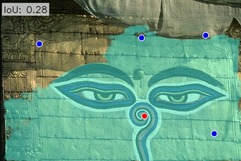}\\ 
Ours & \includegraphics[width=.16\linewidth]{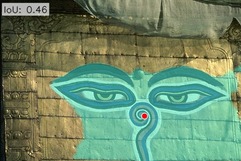}   & \includegraphics[width=.16\linewidth]{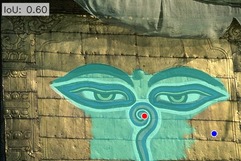}  & \includegraphics[width=.16\linewidth]{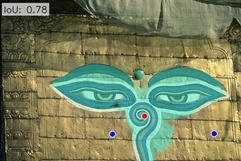}  &   \includegraphics[width=.16\linewidth]{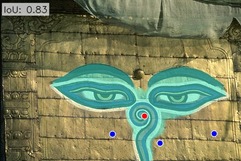} &   \includegraphics[width=.16\linewidth]{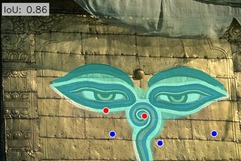} &   \includegraphics[width=.16\linewidth]{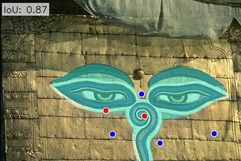}\\

\end{tabular}
\caption{Qualitative comparison between our baseline (BL) and our model across the first 6 clicks. Positive clicks are in red, negative in blue, object masks are highlighted in cyan. Images are from the Berkeley Dataset.}
\label{fig:Berkeleycompare_iter_clicks}
\end{figure*}

In Figure~\ref{fig:Grabcutcompare_iter_clicks} and
\ref{fig:Berkeleycompare_iter_clicks} we compare the predictions of our baseline
and full algorithms for six clicks on images from the GrabCut and Berkeley
datasets. Note that the baseline approach sometimes fails to recover from poor
initial guesses, whereas our iteratively trained network is better at correcting
with each click.

\begin{figure}[p]
\setlength{\tabcolsep}{1pt}
 \centering
\begin{tabular}{ccccc}
BL & \includegraphics[width=.24\linewidth]{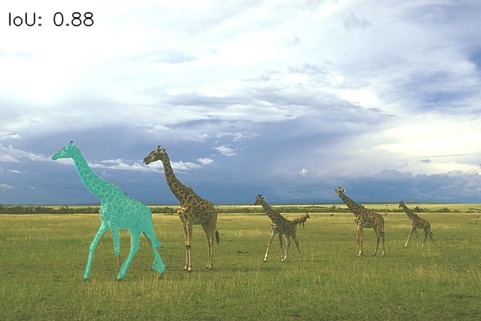} & \includegraphics[width=.24\linewidth]{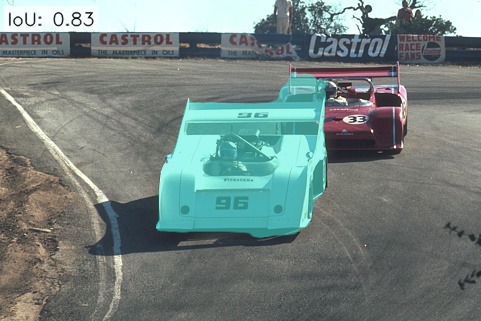} & \includegraphics[width=.24\linewidth]{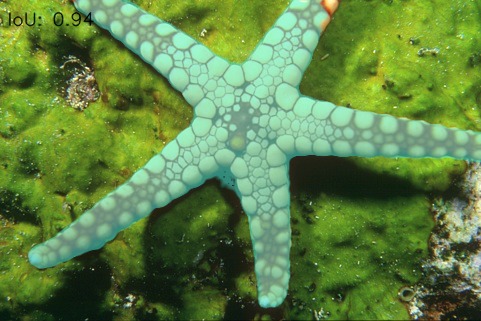} & \includegraphics[width=.24\linewidth]{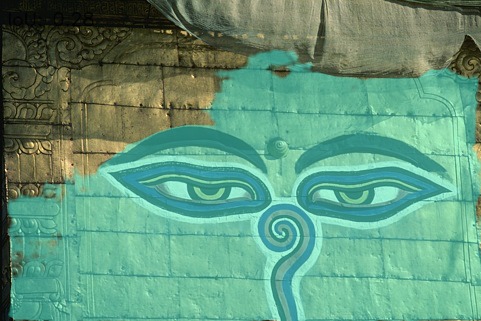} \\
Ours  & \includegraphics[width=.24\linewidth]{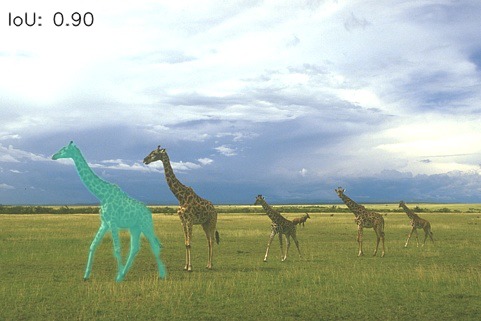} &
\includegraphics[width=.24\linewidth]{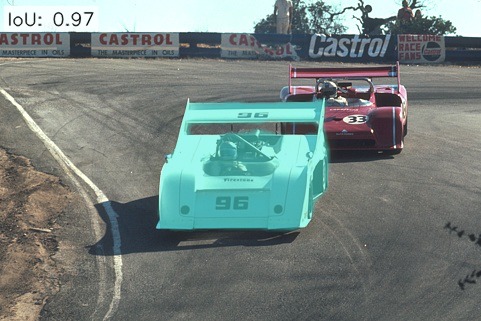}  &  \includegraphics[width=.24\linewidth]{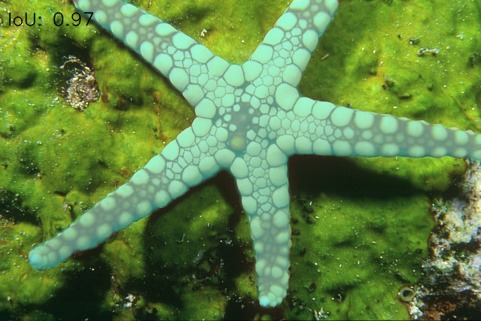} &
\includegraphics[width=.24\linewidth]{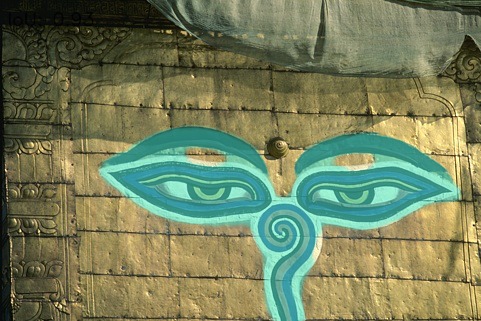}
\end{tabular}
\caption{Qualitative comparison between our baseline and our model after 20 clicks. Object masks are highlighted in cyan.}\label{fig:compare_at_20}
\end{figure}

\noindent In Figure~\ref{fig:compare_at_20} we see that, after 20 clicks, both
models roughly select the object, but our final model is visibly more precise in
these instances.

\begin{figure}[p]
\setlength{\tabcolsep}{1pt}
 \centering
\begin{tabular}{cccc}
\includegraphics[height=.16\linewidth, width=.24\linewidth]{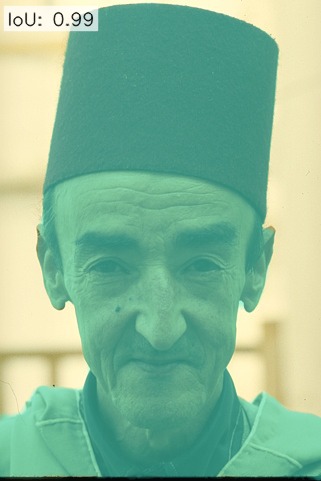} & \includegraphics[height=.16\linewidth,width=.24\linewidth]{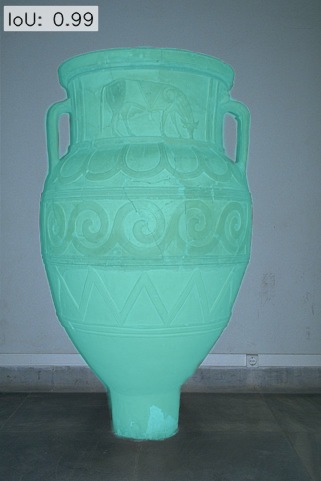} & \includegraphics[height=.16\linewidth,width=.24\linewidth]{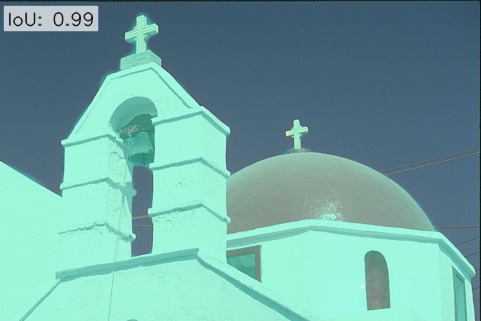}& \includegraphics[height=.16\linewidth,width=.24\linewidth]{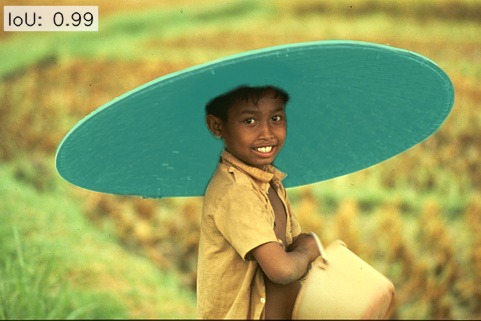} \\
\includegraphics[height=.16\linewidth,
  width=.24\linewidth]{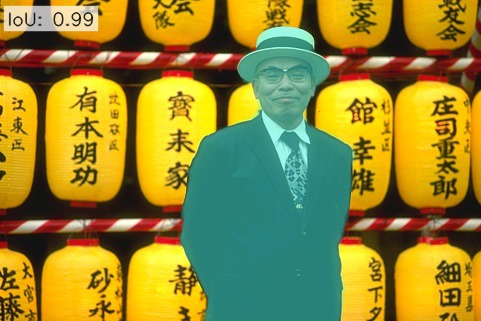} &
\includegraphics[height=.16\linewidth,width=.24\linewidth]{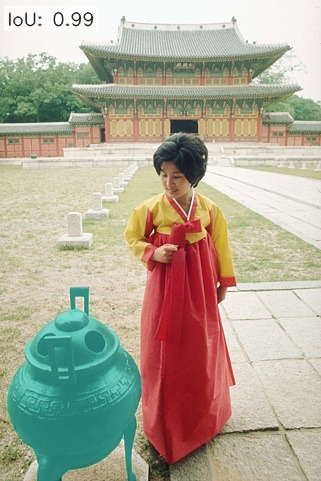}
&
\includegraphics[height=.16\linewidth,width=.24\linewidth]{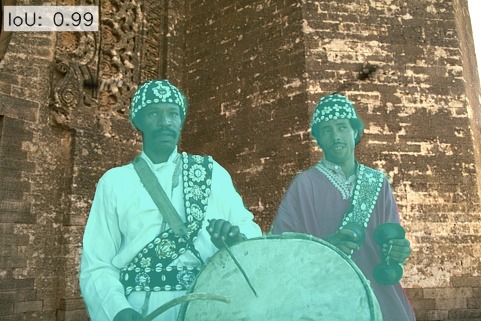}&
\includegraphics[height=.16\linewidth,width=.24\linewidth]{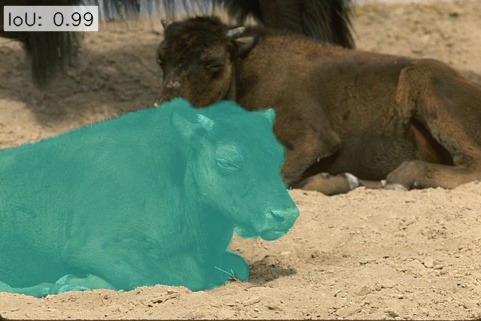}
\\\includegraphics[height=.16\linewidth,width=.24\linewidth]{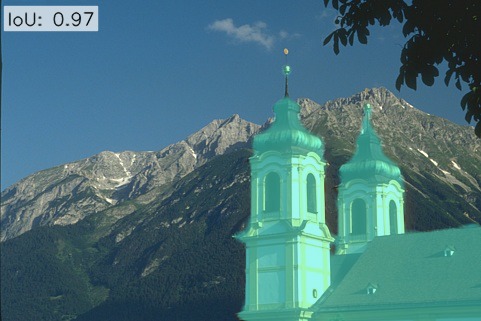} &
\includegraphics[height=.16\linewidth,
  width=.24\linewidth]{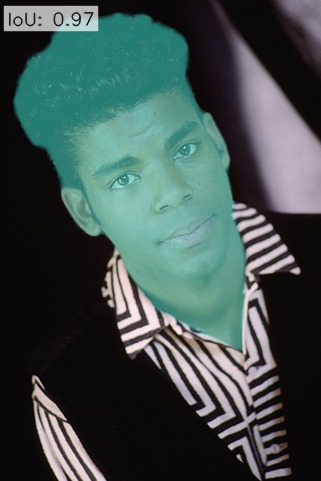} &
\includegraphics[height=.16\linewidth,width=.24\linewidth]{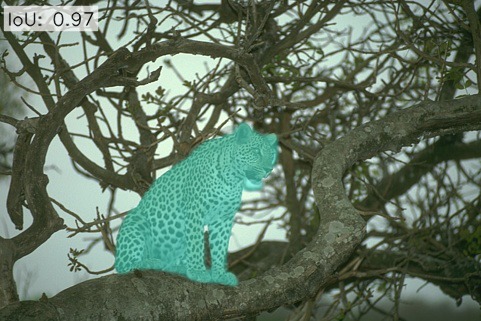}
&
\includegraphics[height=.16\linewidth,width=.24\linewidth]{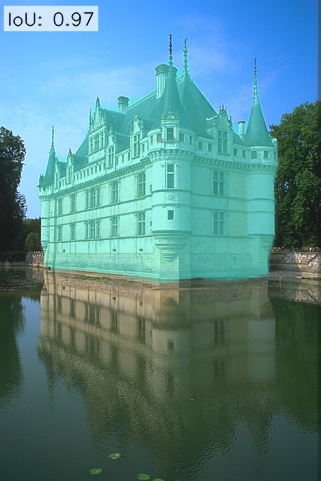}
\\\includegraphics[height=.16\linewidth,width=.24\linewidth]{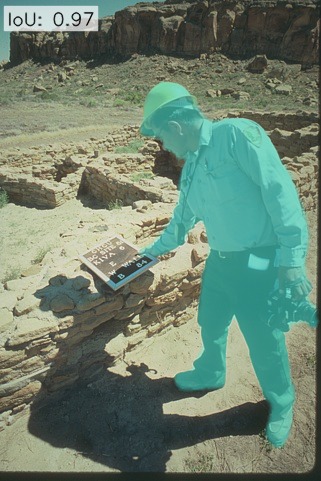}& \includegraphics[height=.16\linewidth,width=.24\linewidth]{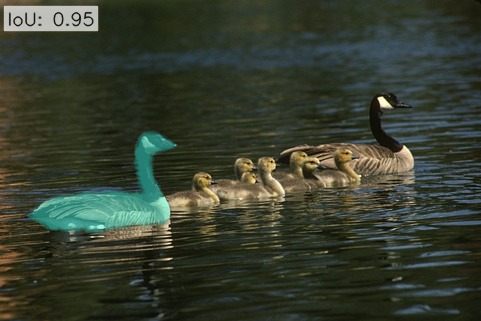} &
\includegraphics[height=.16\linewidth,
  width=.24\linewidth]{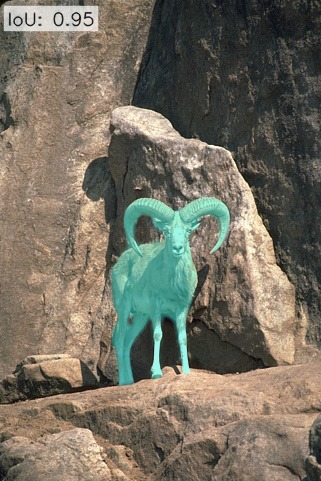} &
\includegraphics[height=.16\linewidth,width=.24\linewidth]{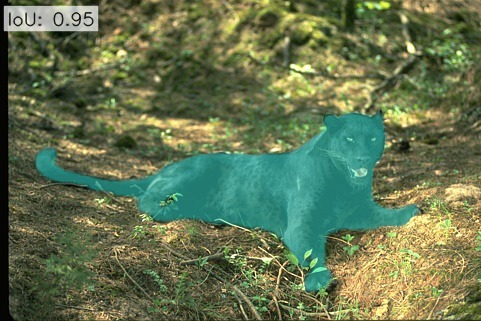}
\\ \includegraphics[height=.16\linewidth,width=.24\linewidth]{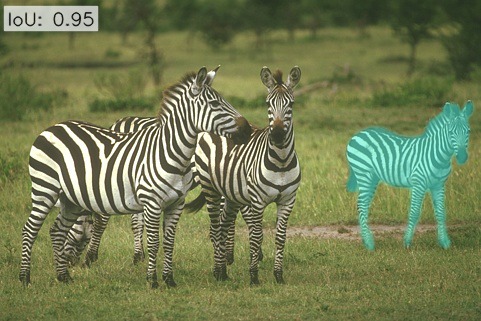}&
\includegraphics[height=.16\linewidth,width=.24\linewidth]{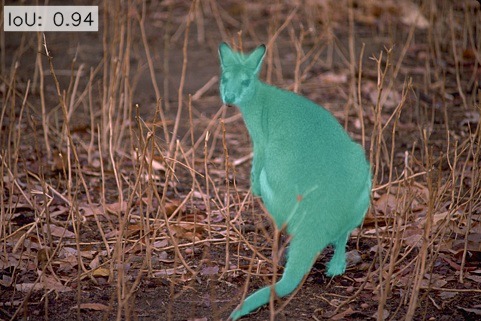}&
\includegraphics[height=.16\linewidth,width=.24\linewidth]{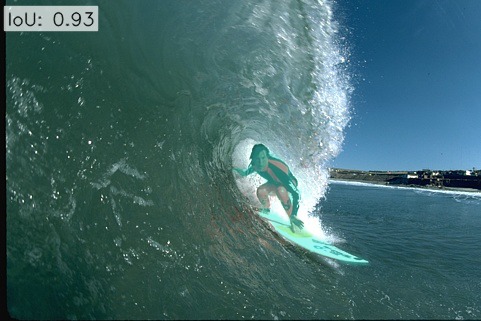}
& 
\includegraphics[height=.16\linewidth,width=.24\linewidth]{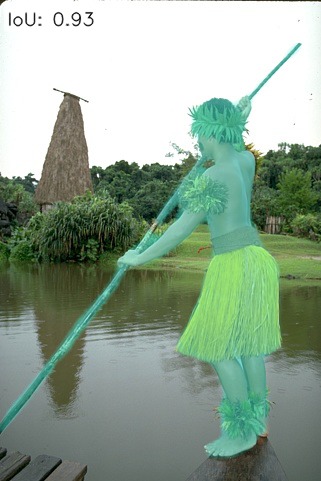} \\
\includegraphics[height=.16\linewidth,width=.24\linewidth]{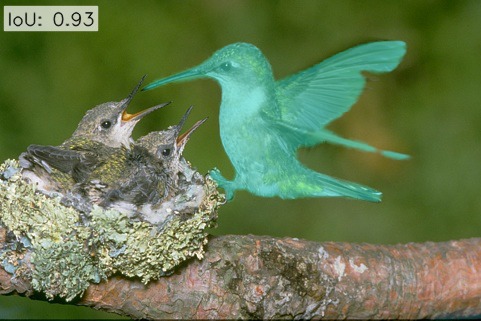}  &  \includegraphics[height=.16\linewidth,width=.24\linewidth]{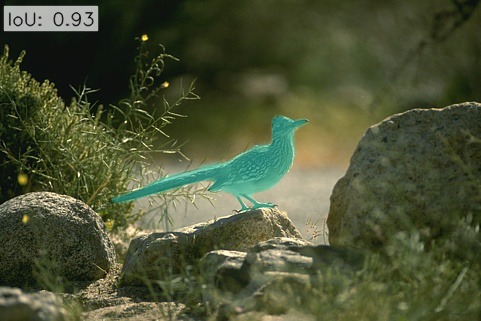} &
\includegraphics[height=.16\linewidth,width=.24\linewidth]{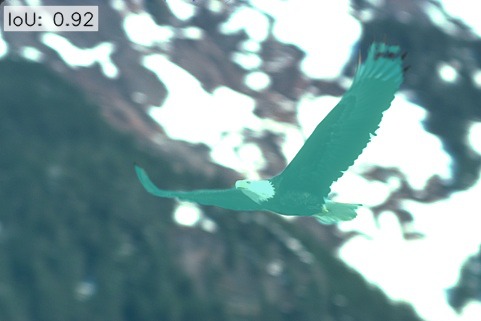} &
\includegraphics[height=.16\linewidth,width=.24\linewidth]{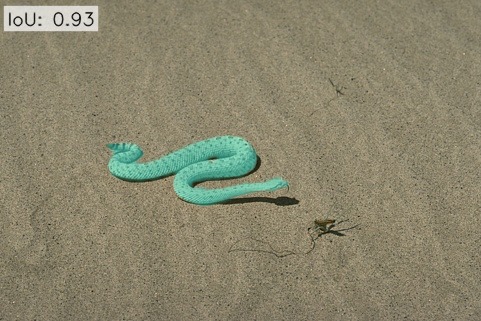}
\\ \includegraphics[height=.16\linewidth,width=.24\linewidth]{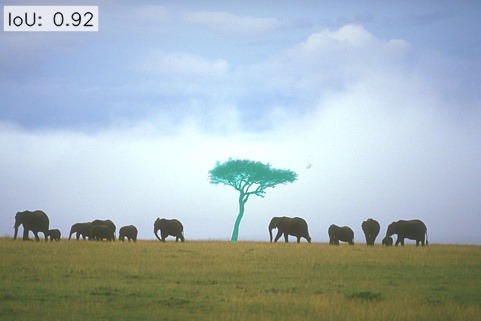}
&
\includegraphics[height=.16\linewidth,width=.24\linewidth]{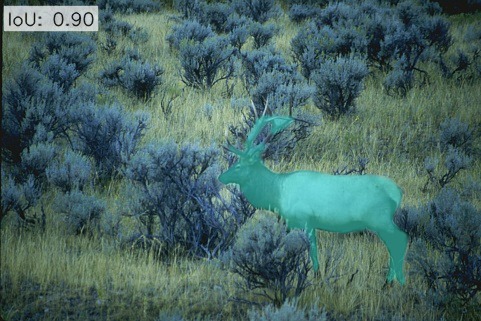}
&
\includegraphics[height=.16\linewidth,width=.24\linewidth]{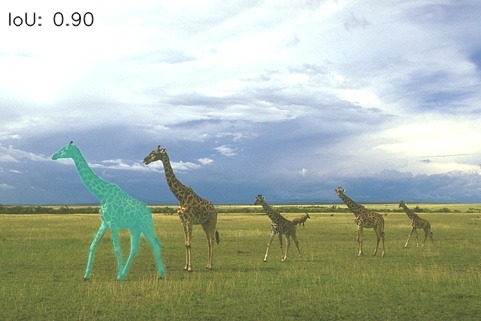}
 &
 \includegraphics[height=.16\linewidth,width=.24\linewidth]{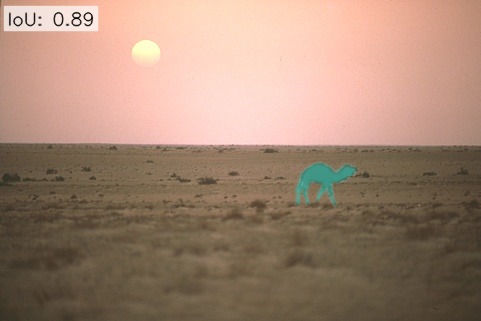} %

\end{tabular}
\caption{Example predictions from our method at 20 clicks on the Berkeley dataset. Examples are ordered from the easiest (top) to the most difficult images (bottom), based on the mean IoU accuracy over the 15-20 click range. Positive clicks are in red, negative in blue, object masks are highlighted in cyan. }\label{fig:BerkFinalDiff}
\end{figure}

To illustrate the behaviour of our method across the whole range of
difficulties, we sampled in Figure~\ref{fig:BerkFinalDiff} images based on
their accuracy at 20 clicks. The images are ranked from easiest (top) to hardest
(bottom).

\section{Training on a Synthetic Dataset}
\label{sec:synthetic}

\begin{figure}[t]
    \centering
    \begin{tabular}{cc}
 \includegraphics[width=.24\textwidth,height=.15\textwidth]{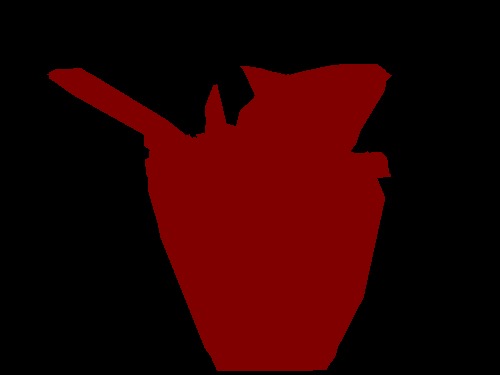}\includegraphics[width=.24\textwidth,height=.15\textwidth]{fig/PoorBoundarySetExamples/SBD_Set_examples/train/2008_000188.jpg}  & 
    \includegraphics[width=.24\textwidth, height=.15\textwidth]{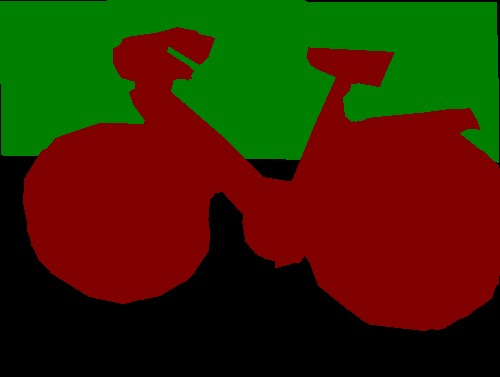}\includegraphics[width=.24\textwidth, height=.15\textwidth]{fig/PoorBoundarySetExamples/SBD_Set_examples/val/2008_000133.jpg}
    \end{tabular}
    \caption{Example of labelling errors in SBD train and validation dataset.}
    \label{fig:dataset:imprecise}
\end{figure}

The two most popular training sets used in interactive segmentation are PASCAL 2012~\cite{pascal-voc-2012,BharathICCV2011} and SBD~\cite{lin2014microsoft}. They contain thousands of images labelled by annotators. However, these datasets are also quite imprecise, which is a problem when training very high-quality segmentation networks. Examples of ground-truth delineations can be found in Figure~\ref{fig:dataset:imprecise}. 
Interestingly, the GrabCut benchmark defines a band of pixel around the object boundary that is excluded when measuring the accuracy. This is not the case with SBD and PASCAL, for which every pixel is evaluated. This partly explains why state of the art techniques achieve higher levels of accuracy on GrabCut.

We can argue that whatever is drawn by the annotators should be considered as
ground truth, and that there is no such thing as an incorrect labelling. After
all, interactive segmentation algorithms are meant to assist the user achieve
whatever arbitrary delineation they want. However, the labelled masks on these
datasets would not be good enough in high-end image or video editing
applications. These existing datasets are thus not representative of our final
application.

We therefore propose to explore how training our network on an accurate and
consistently labelled synthetic dataset would impact our overall performance.

\subsection{Synthetic Dataset Construction}

\begin{figure*}[p]
\setlength{\tabcolsep}{0.1em}
    \includegraphics[width=\linewidth]{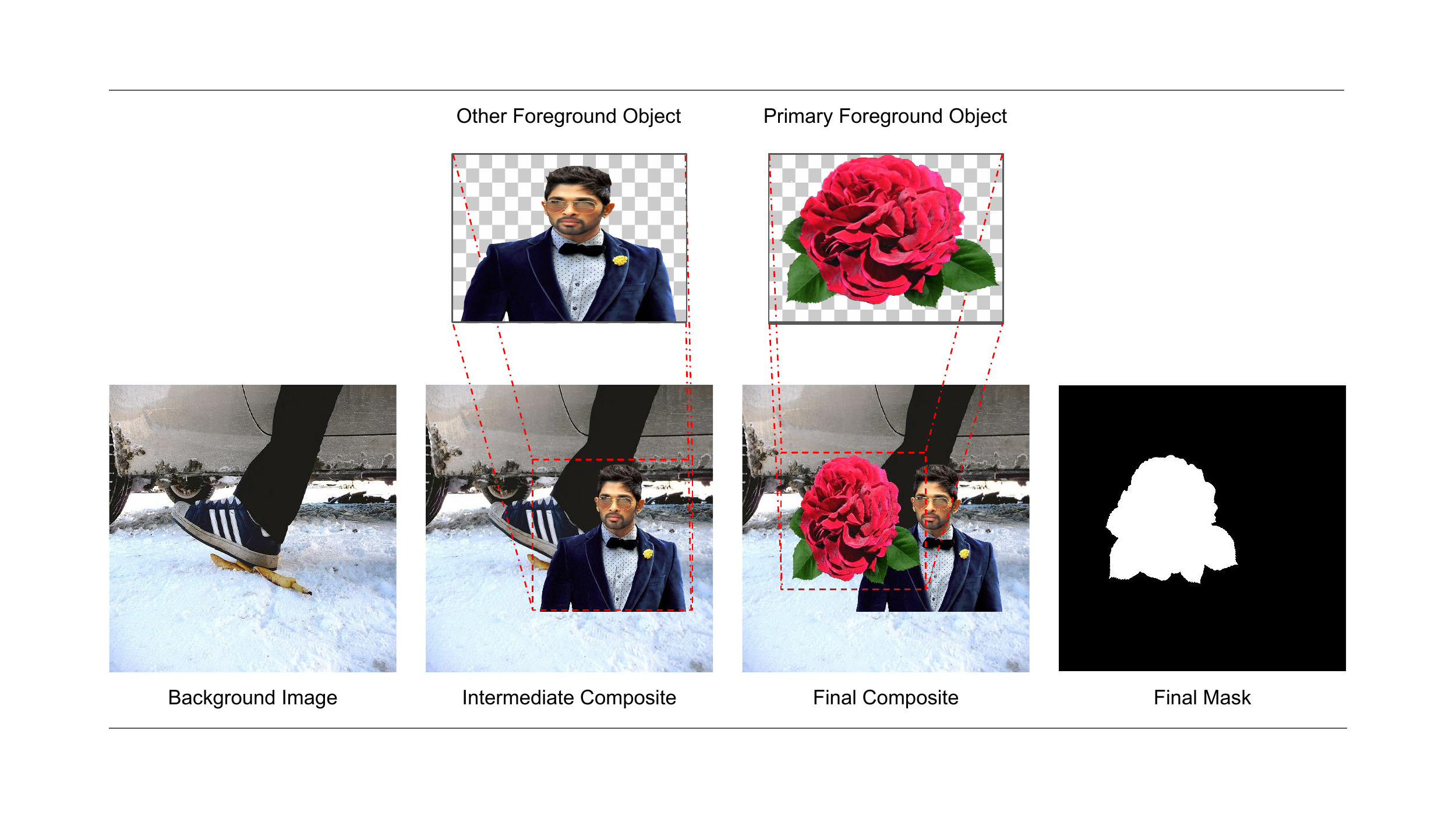}\vspace{-2em}
    \caption{{\bf Creating an image of the synthetic dataset.}}
    \label{fig:synthetic:construction}
\end{figure*}

\begin{figure*}[p]
\setlength{\tabcolsep}{0.1em}
    \begin{tabular}{cccccccc}
    \includegraphics[width=.122\linewidth]{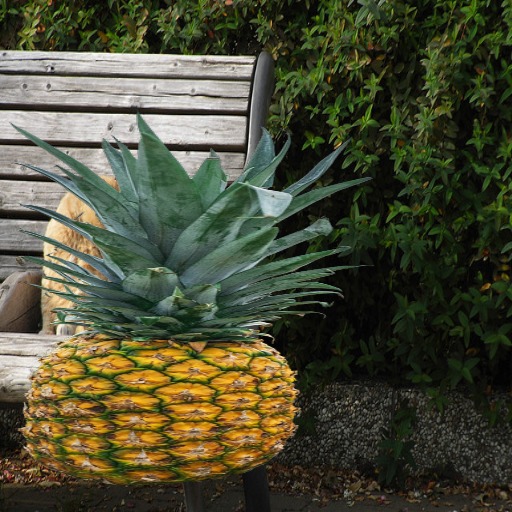} &
    \includegraphics[width=.122\linewidth]{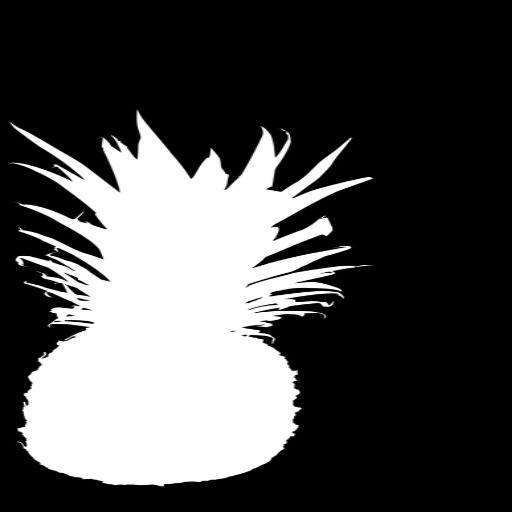} &
    \includegraphics[width=.122\linewidth]{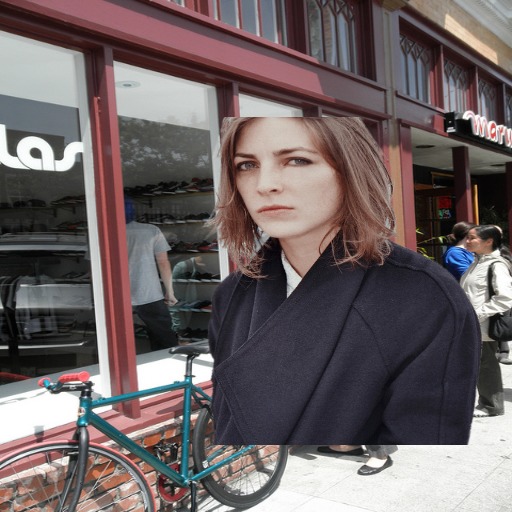} &
    \includegraphics[width=.122\linewidth]{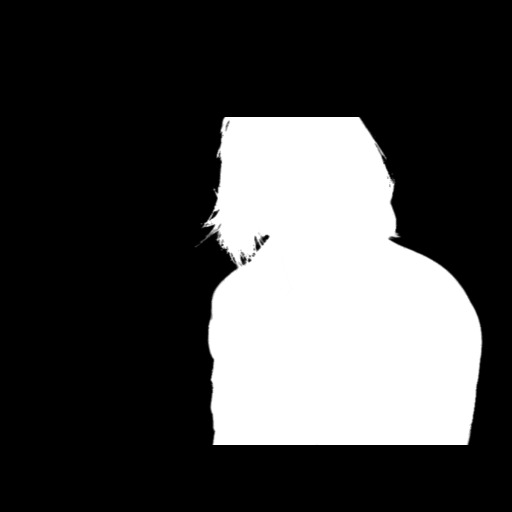} &
    \includegraphics[width=.122\linewidth]{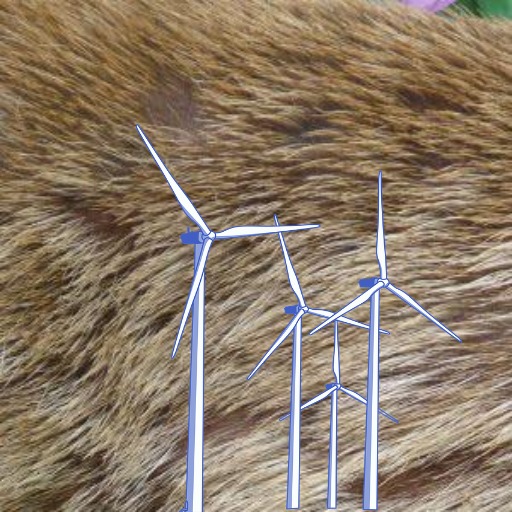} & 
    \includegraphics[width=.122\linewidth]{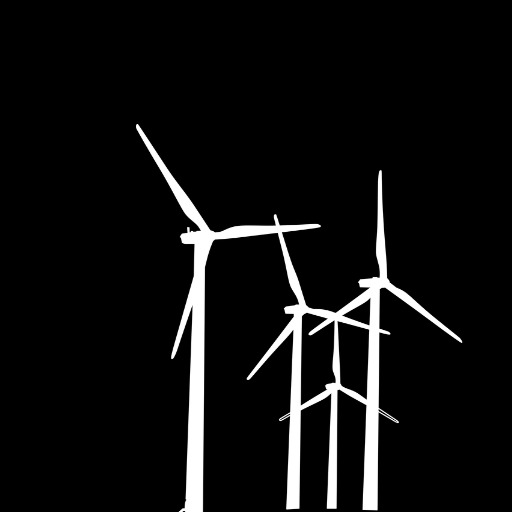} &
    \includegraphics[width=.122\linewidth]{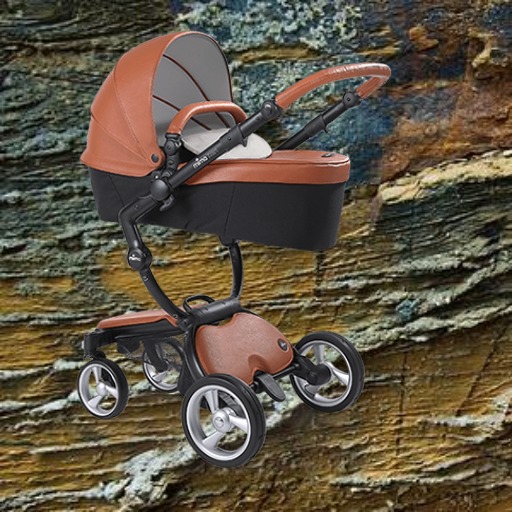} &
    \includegraphics[width=.122\linewidth]{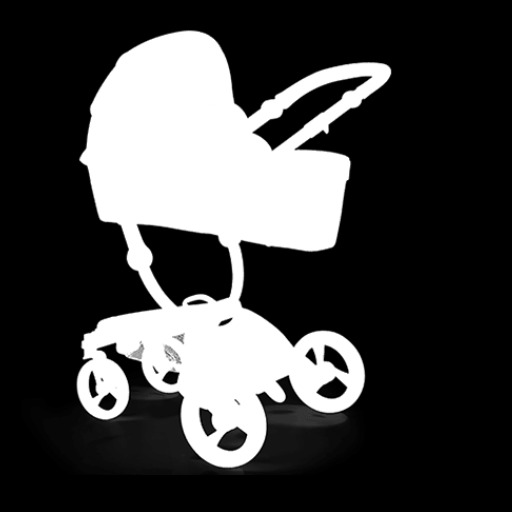} 
    \\
    \includegraphics[width=.122\linewidth]{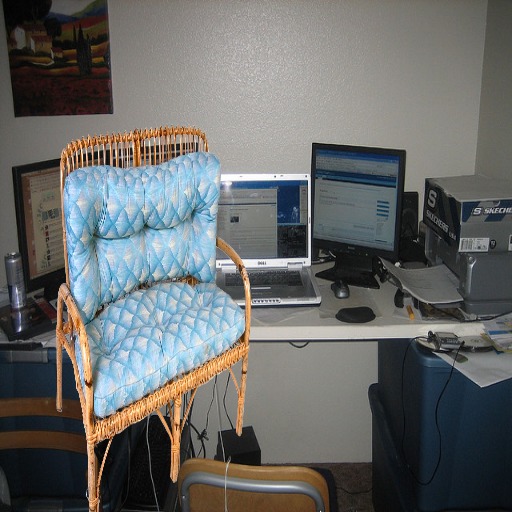} & 
    \includegraphics[width=.122\linewidth]{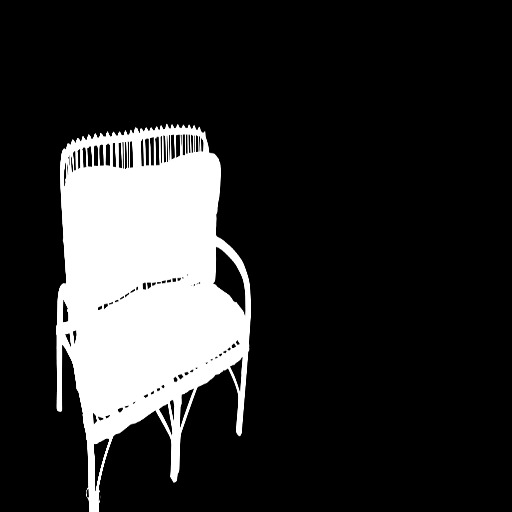} &
    \includegraphics[width=.122\linewidth]{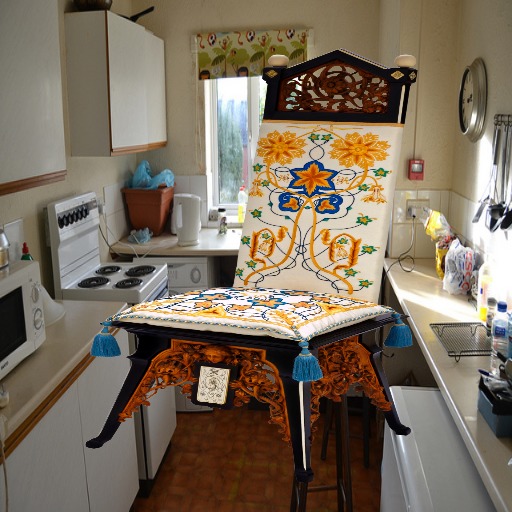} &
    \includegraphics[width=.122\linewidth]{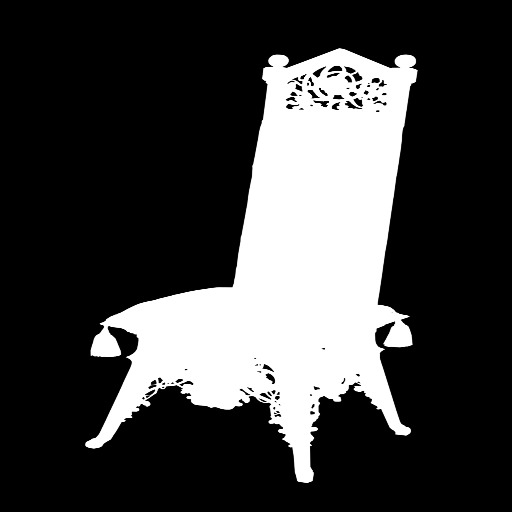} &
    \includegraphics[width=.122\linewidth]{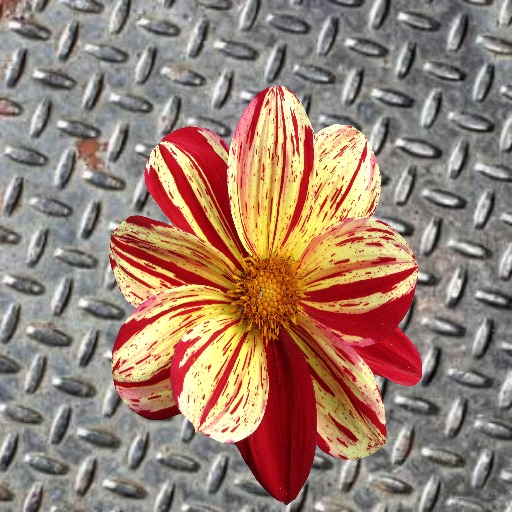} &
    \includegraphics[width=.122\linewidth]{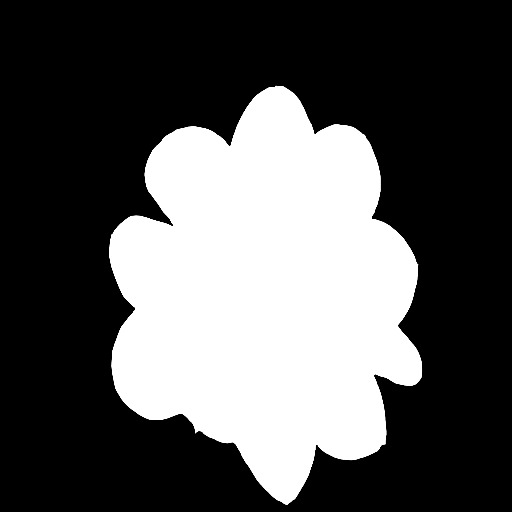} & 
    \includegraphics[width=.122\linewidth]{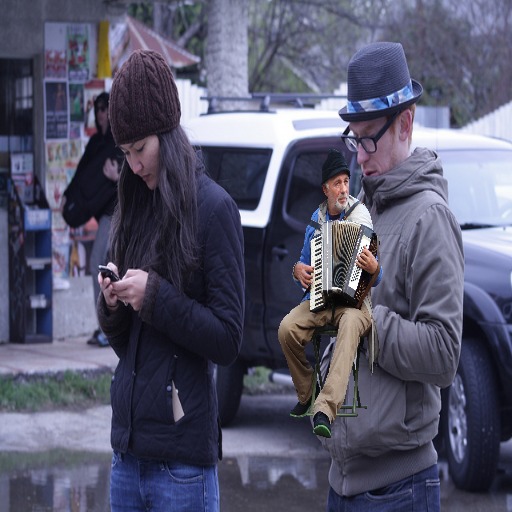} &
    \includegraphics[width=.122\linewidth]{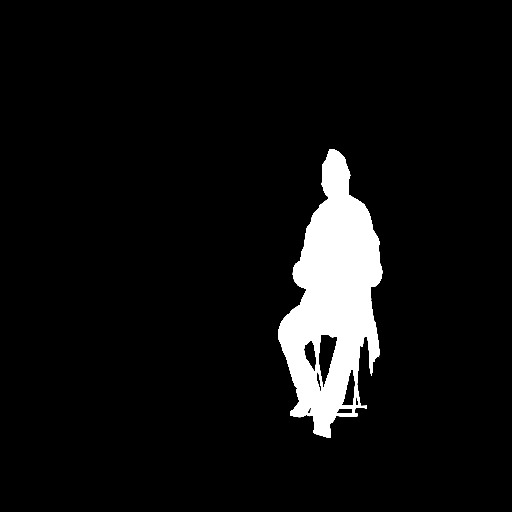}
    \\
    \includegraphics[width=.122\linewidth]{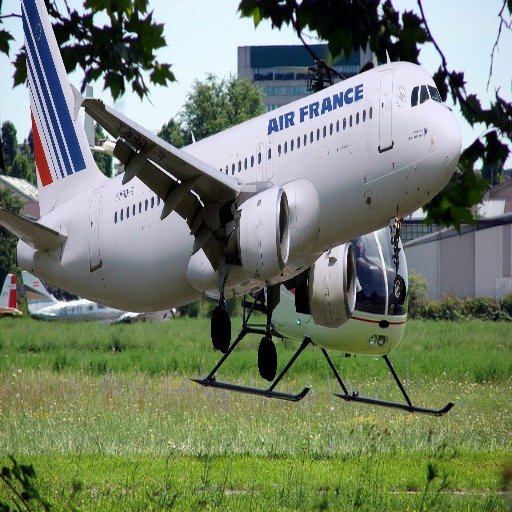} & 
    \includegraphics[width=.122\linewidth]{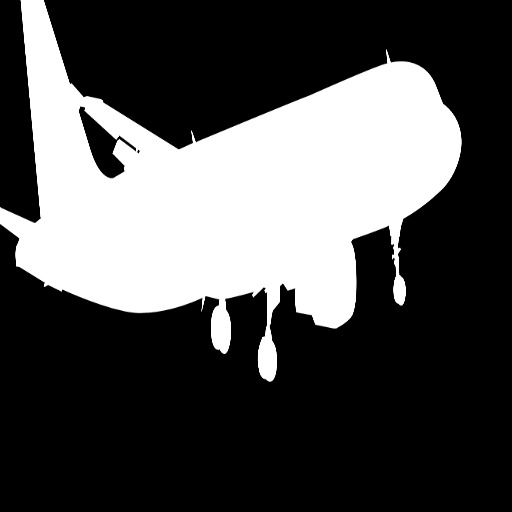} &
    \includegraphics[width=.122\linewidth]{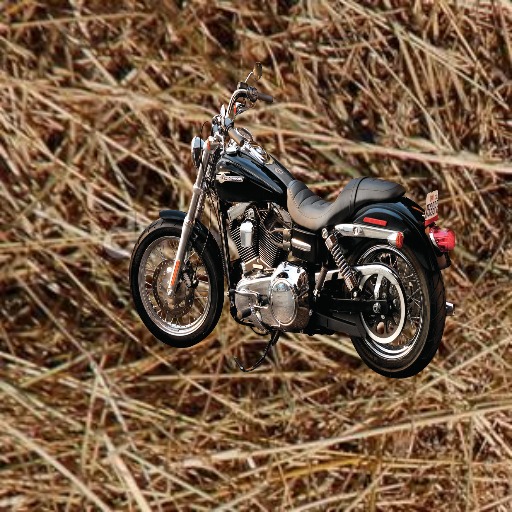} &
    \includegraphics[width=.122\linewidth]{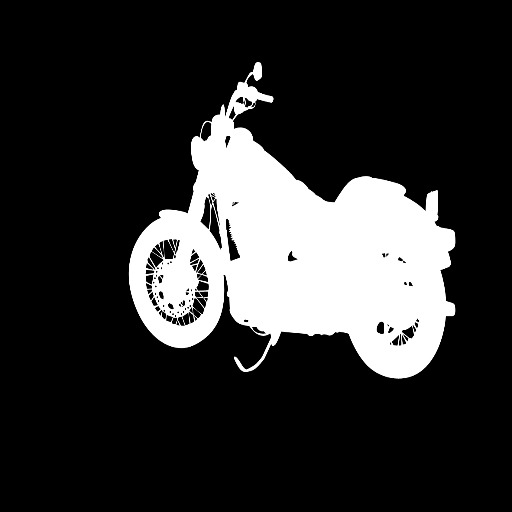} &
    \includegraphics[width=.122\linewidth]{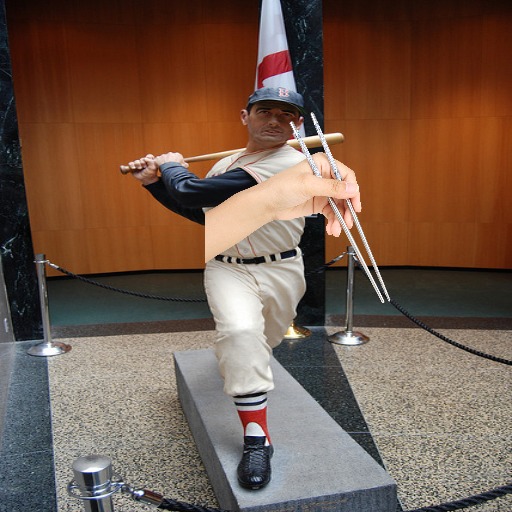} &
    \includegraphics[width=.122\linewidth]{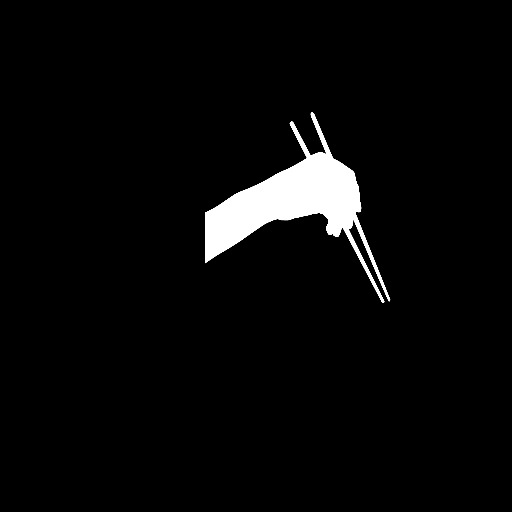} & 
    \includegraphics[width=.122\linewidth]{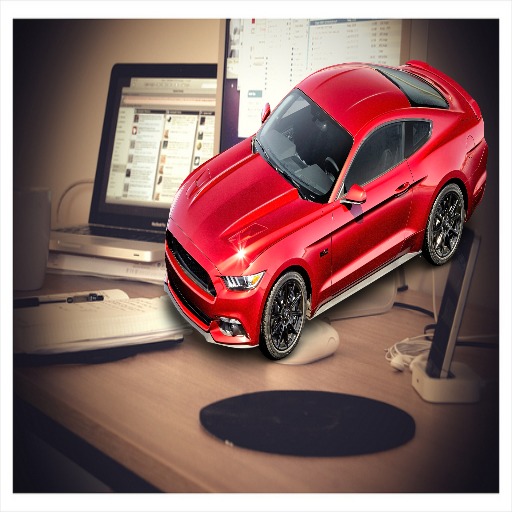} &
    \includegraphics[width=.122\linewidth]{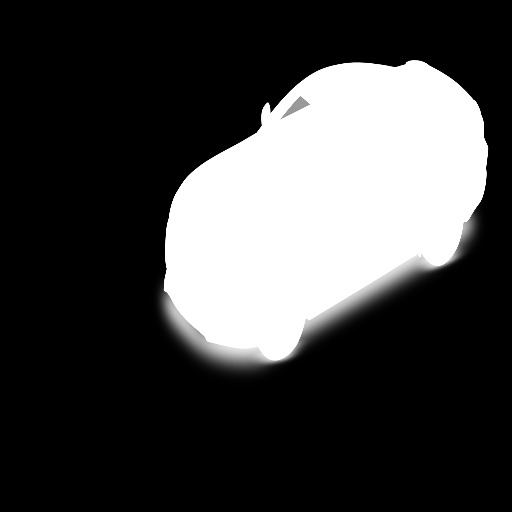}
    \\
    \includegraphics[width=.122\linewidth]{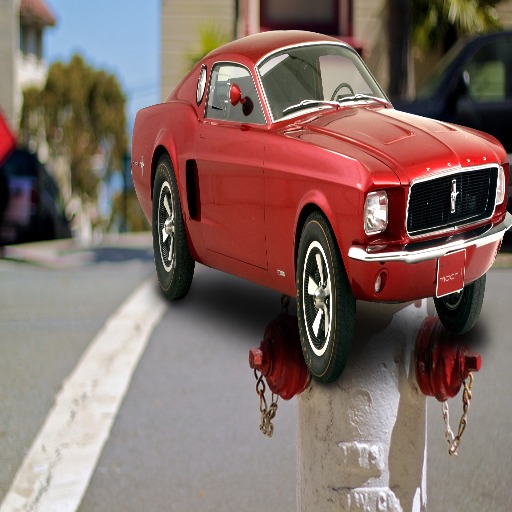} & 
    \includegraphics[width=.122\linewidth]{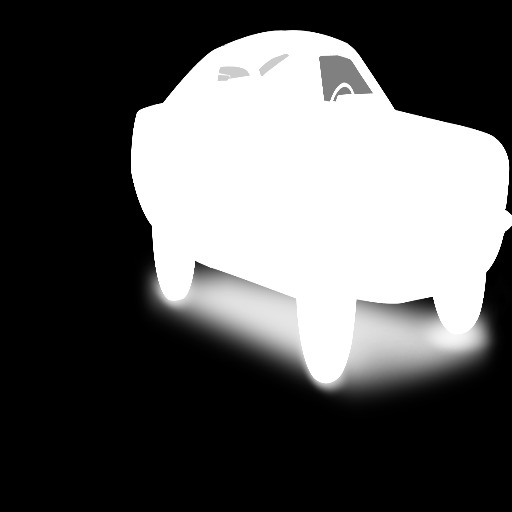} &
    \includegraphics[width=.122\linewidth]{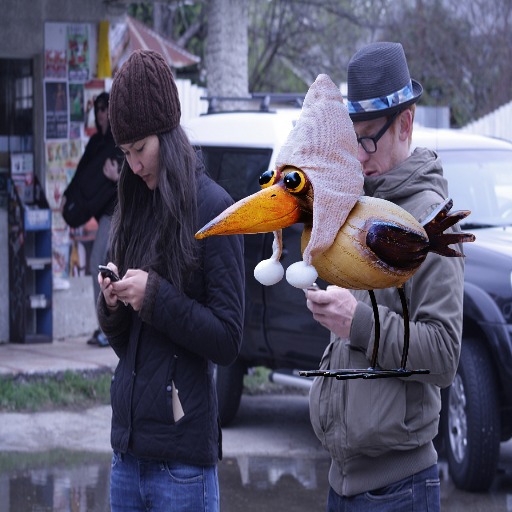}  &
    \includegraphics[width=.122\linewidth]{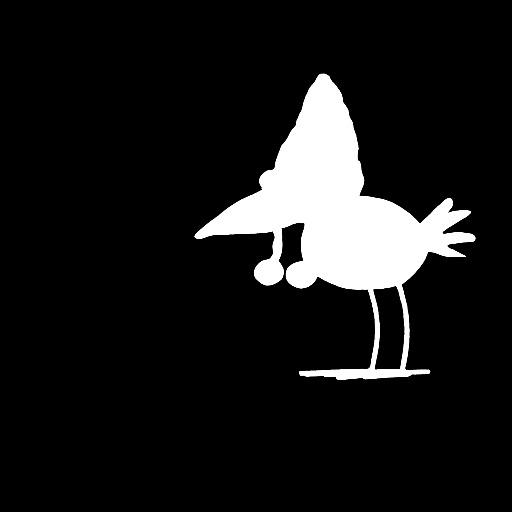} &
    \includegraphics[width=.122\linewidth]{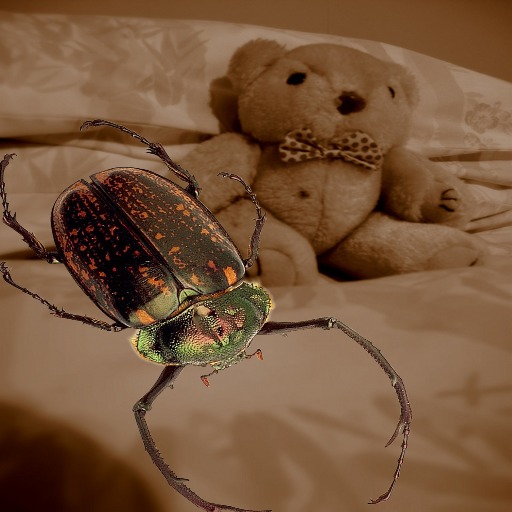} & 
    \includegraphics[width=.122\linewidth]{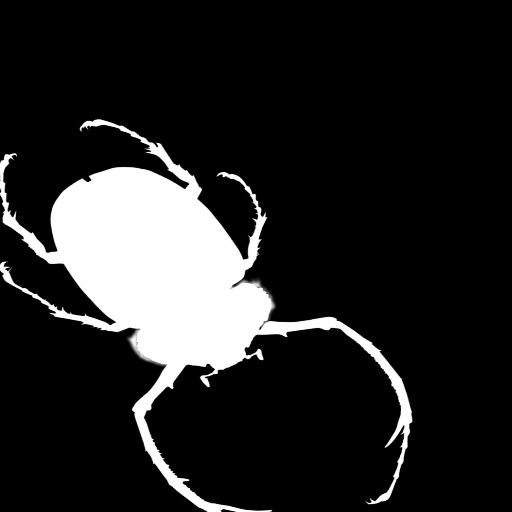} &
    \includegraphics[width=.122\linewidth]{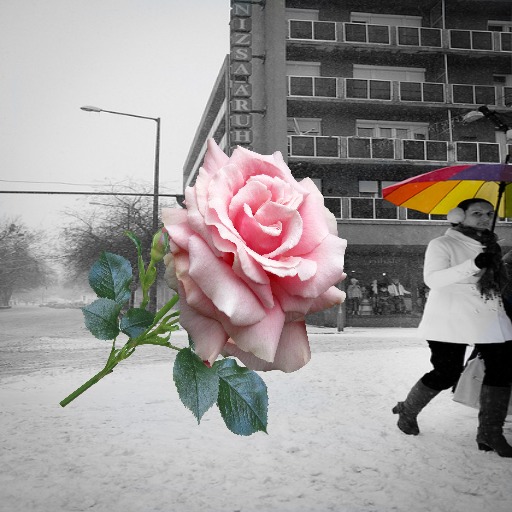} &
    \includegraphics[width=.122\linewidth]{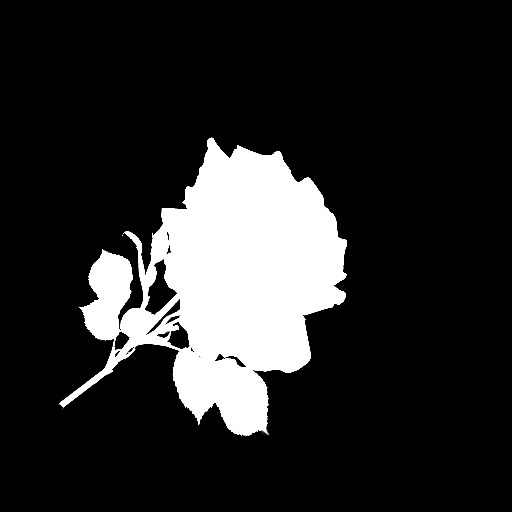} \\  
    \includegraphics[width=.122\linewidth]{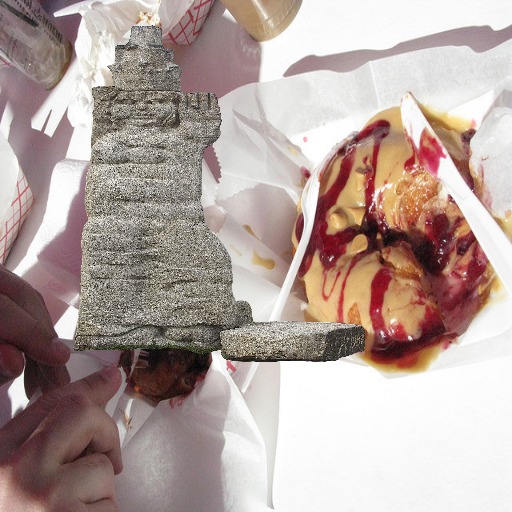} &
    \includegraphics[width=.122\linewidth]{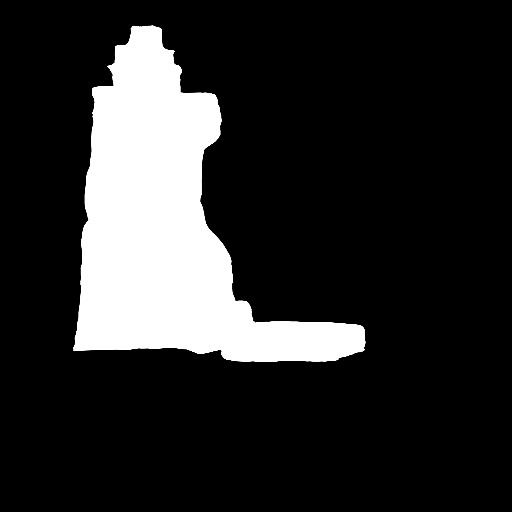} &
     \includegraphics[width=.122\linewidth]{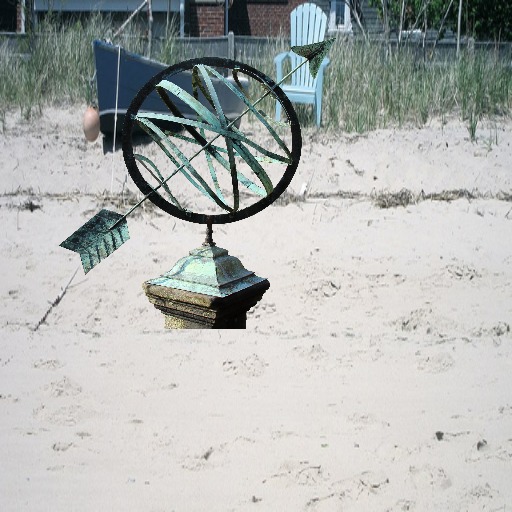}& 
    \includegraphics[width=.122\linewidth]{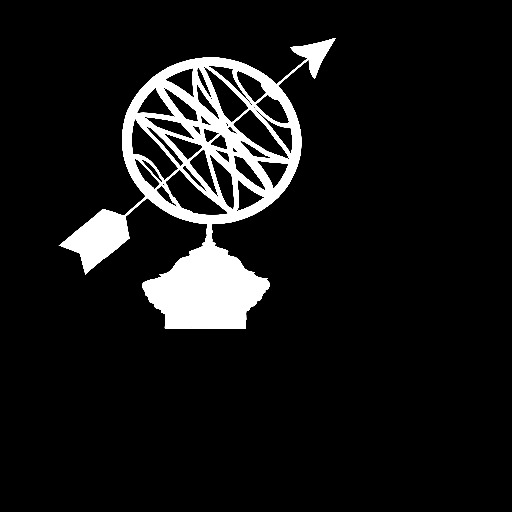} &
    \includegraphics[width=.122\linewidth]{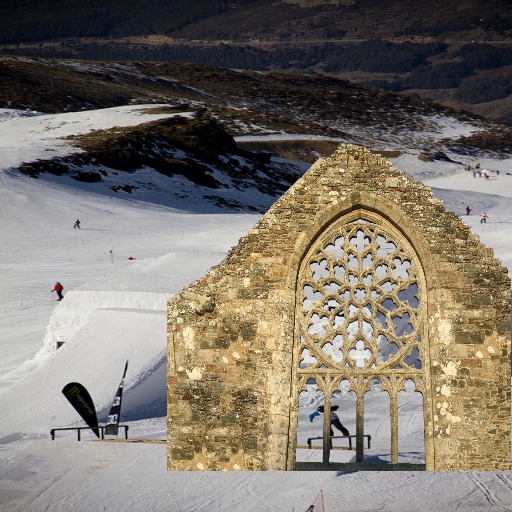} & 
    \includegraphics[width=.122\linewidth]{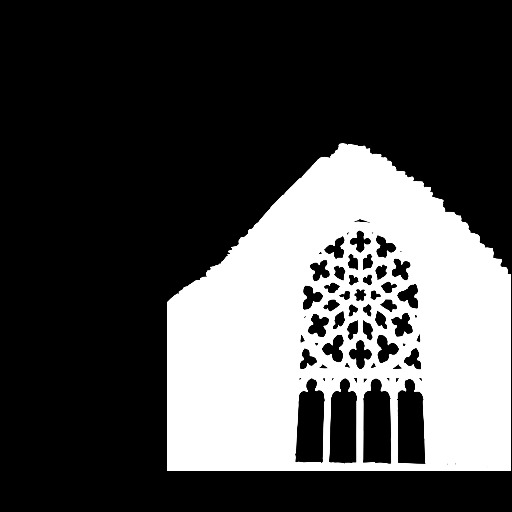} &
    \includegraphics[width=.122\linewidth]{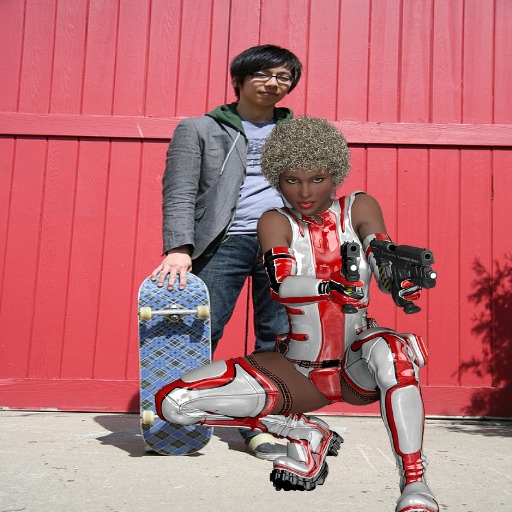} &
    \includegraphics[width=.122\linewidth]{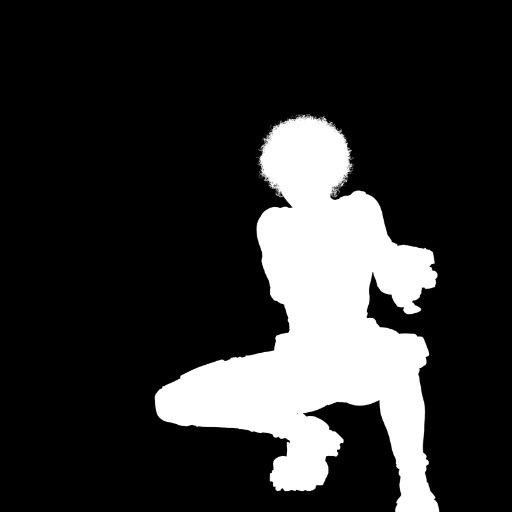} \\
    \end{tabular}
    \caption{{\bf Synthetic Dataset Examples}: Variety of training images and their corresponding masks, generated from our Synthetic Dataset.}
    \label{fig:synthetic:examples}
\end{figure*}

We constructed our synthetic training dataset in a similar way to the Deep Image
Matting dataset from \citet{Xu_2017_CVPR}. We sourced 15,000 images of objects
with foreground colours and transparencies from the Internet. As illustrated in
Figure~\ref{fig:synthetic:construction}, the images are synthesised on the fly
during training by compositing one of these foreground objects onto a randomly
sampled background. The background images come from
MS-COCO~\cite{lin2014microsoft} and ETHZ Synthesizability
\cite{dai:synthesizability} datasets. The binary masks are obtained by
thresholding the alpha matte at 0.5.  A few examples of such composite training
images are shown in Figure~\ref{fig:synthetic:examples}.


\subsection{Synthetic Dataset Training Results}

We propose to analyse the following three training scenarios:
\begin{enumerate}%
\item our full model (\oursItertrainEncoderFeedbackGF), trained on SBD as proposed in the previous sections
\item our full model (\oursItertrainEncoderFeedbackGFPT), solely trained on the synthetic dataset for 10 epochs at fixed learning rate $10^{-5}$.
\item our full model (\oursItertrainEncoderFeedbackGFPTFT),  finetuning the previous model on SBD for 5 epochs with learning rate $5\times 10^{-6}$.
\end{enumerate}

\begin{figure*}[p]
\setlength{\tabcolsep}{0.1em}
    \begin{tabular}{ccccc}
    Image & Ground Truth &  SBD \cite{BharathICCV2011} &   Synthetic & Synthetic+ SBD \\ [0.5ex] 
    \hline
   \includegraphics[height=.13\linewidth, width=.193\linewidth]{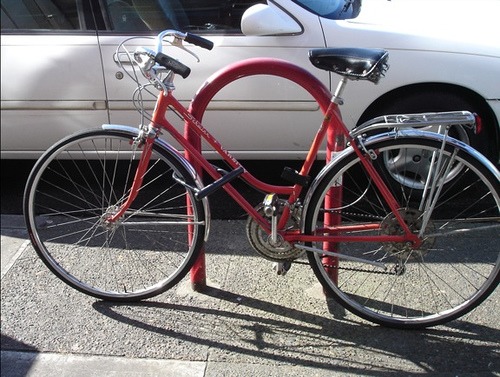} &
    \includegraphics[height=.13\linewidth, width=.193\linewidth]{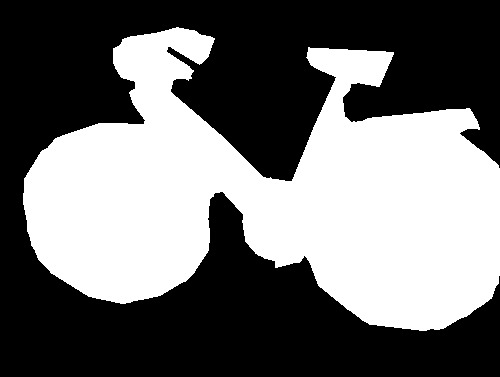} &
    \includegraphics[height=.13\linewidth, width=.193\linewidth]{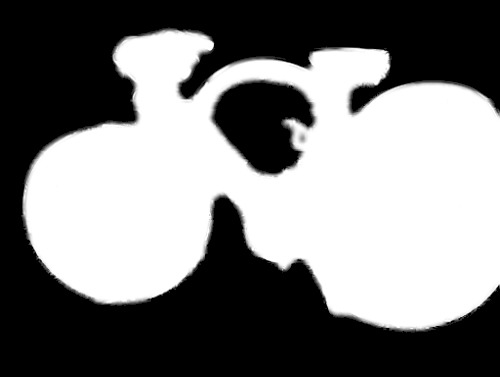} &
    \includegraphics[height=.13\linewidth, width=.193\linewidth]{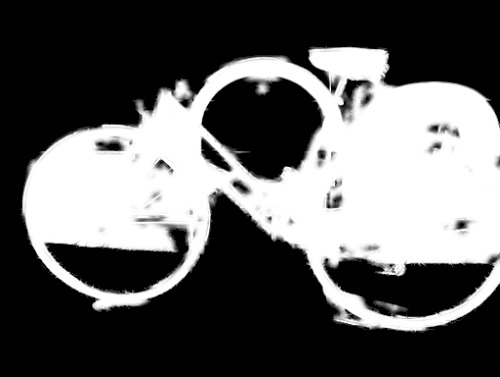} & 
    \includegraphics[height=.13\linewidth, width=.193\linewidth]{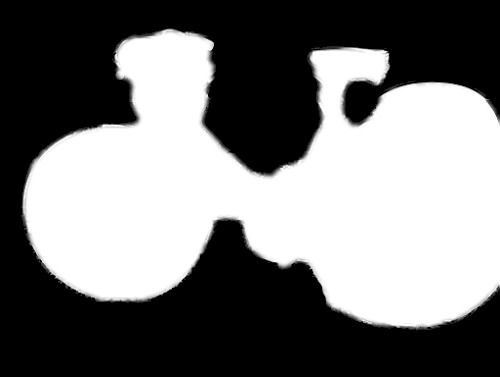} \\
      \includegraphics[height=.13\linewidth, width=.193\linewidth]{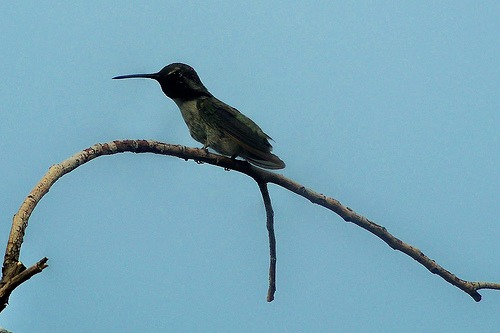} &
    \includegraphics[height=.13\linewidth, width=.193\linewidth]{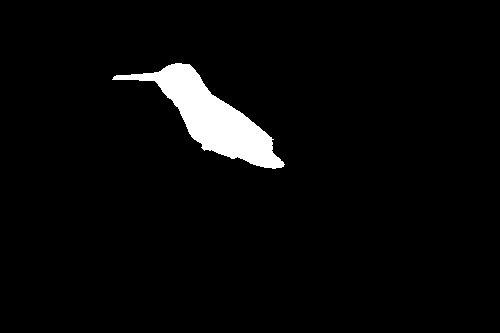} &
    \includegraphics[height=.13\linewidth, width=.193\linewidth]{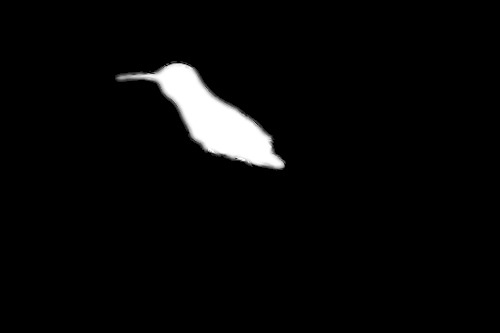} &
    \includegraphics[height=.13\linewidth, width=.193\linewidth]{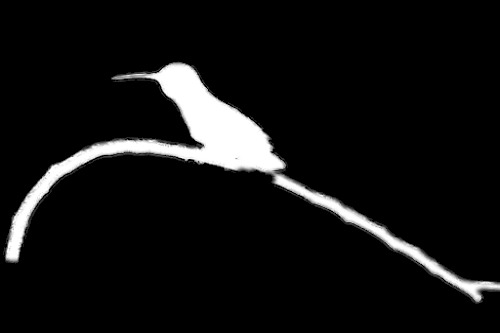} & 
    \includegraphics[height=.13\linewidth, width=.193\linewidth]{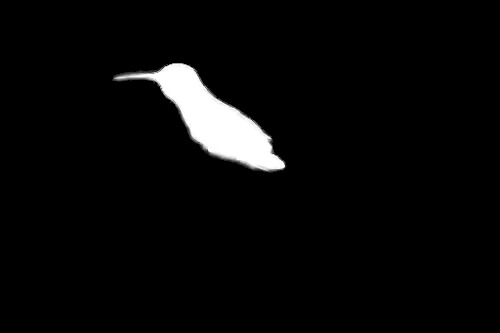} \\
  \includegraphics[height=.13\linewidth, width=.193\linewidth]{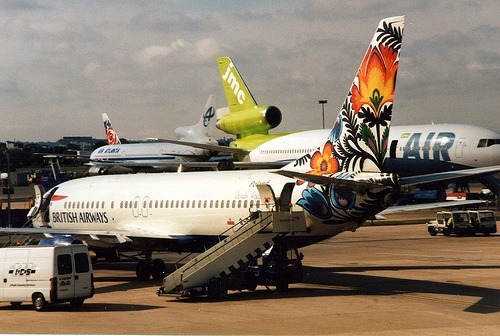} &
    \includegraphics[height=.13\linewidth, width=.193\linewidth]{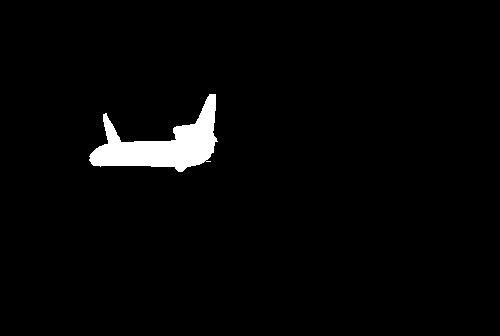} &
    \includegraphics[height=.13\linewidth, width=.193\linewidth]{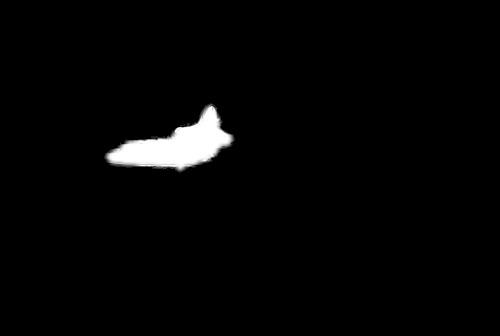} &
    \includegraphics[height=.13\linewidth, width=.193\linewidth]{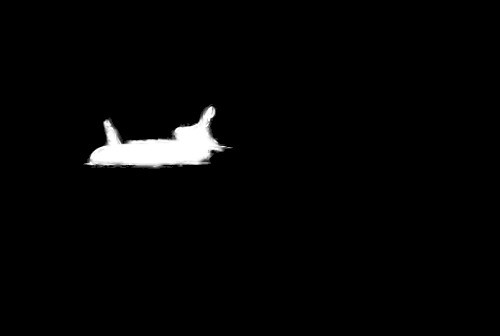} & 
    \includegraphics[height=.13\linewidth, width=.193\linewidth]{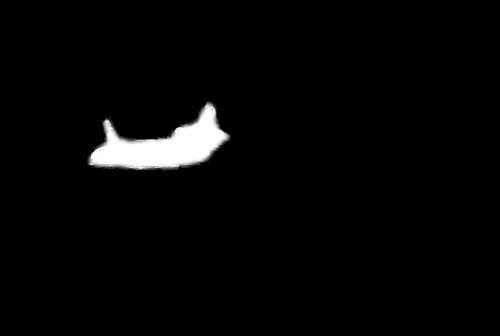} \\
    \end{tabular}
    \caption{{\bf Segmentation examples at 3 clicks, for models trained with SBD, trained with the synthetic dataset alone, and trained with the synthetic set and finetuned on SBD}: Images are from the SBD validation set. }
    \label{fig:synthetic:comparisonsLow}
\end{figure*}

\begin{figure*}[p]
\setlength{\tabcolsep}{0.1em}
    \begin{tabular}{ccccc}
    Image & Ground Truth & SBD~\cite{BharathICCV2011} &  Synthetic & Synthetic + SBD \\ [0.5ex] 
    \hline
    \includegraphics[height=.13\linewidth, width=.193\linewidth]{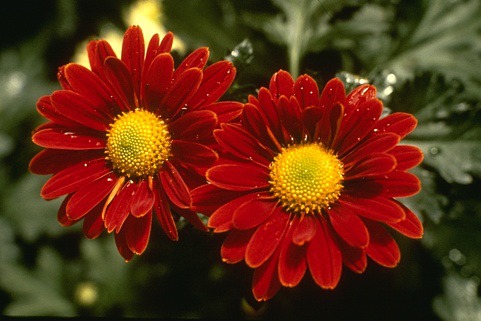} &
    \includegraphics[height=.13\linewidth, width=.193\linewidth]{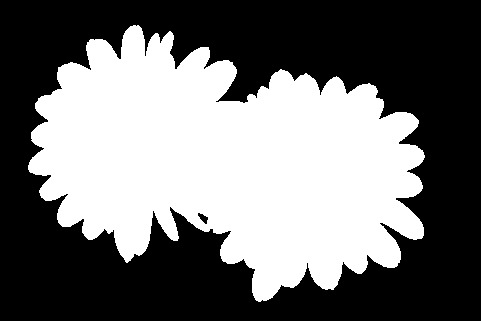} &
    \includegraphics[height=.13\linewidth, width=.193\linewidth]{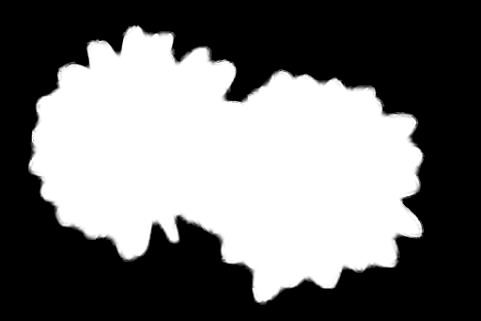} &  
    \includegraphics[height=.13\linewidth, width=.193\linewidth]{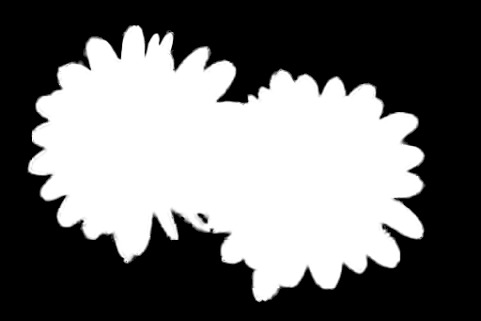} & 
    \includegraphics[height=.13\linewidth, width=.193\linewidth]{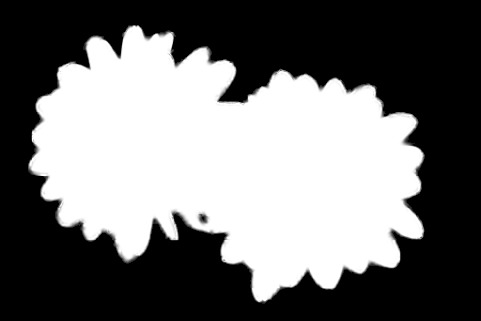} \\
    \includegraphics[height=.13\linewidth, width=.193\linewidth]{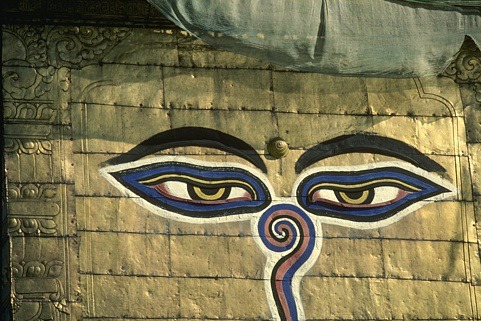} &
    \includegraphics[height=.13\linewidth, width=.193\linewidth]{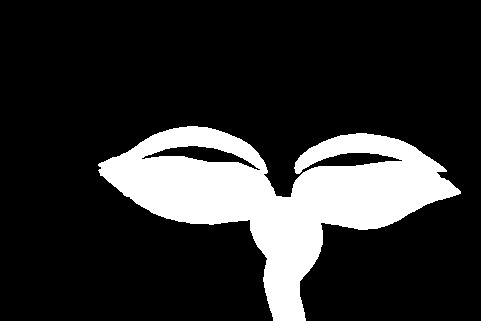} &
    \includegraphics[height=.13\linewidth, width=.193\linewidth]{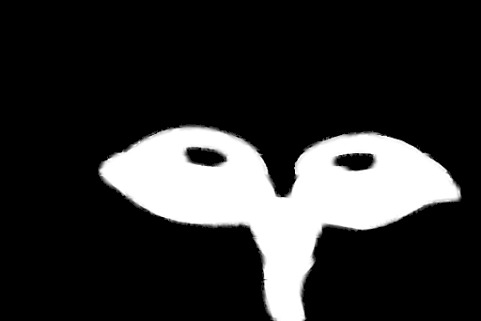} &  
    \includegraphics[height=.13\linewidth, width=.193\linewidth]{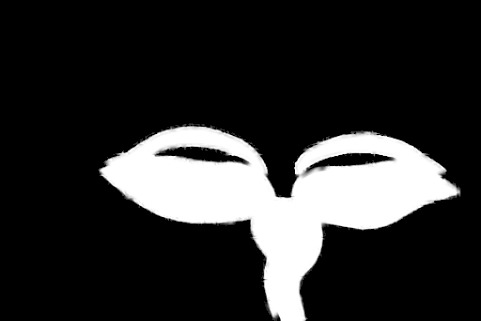} & 
    \includegraphics[height=.13\linewidth, width=.193\linewidth]{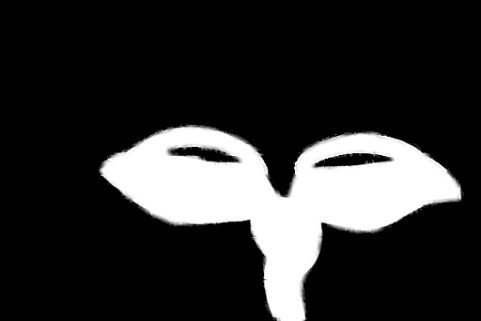} \\
    \includegraphics[height=.13\linewidth, width=.193\linewidth]{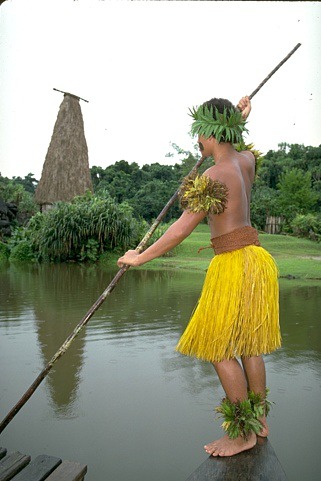} &
    \includegraphics[height=.13\linewidth, width=.193\linewidth]{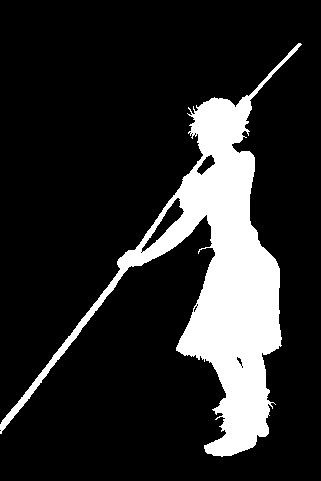} &
    \includegraphics[height=.13\linewidth, width=.193\linewidth]{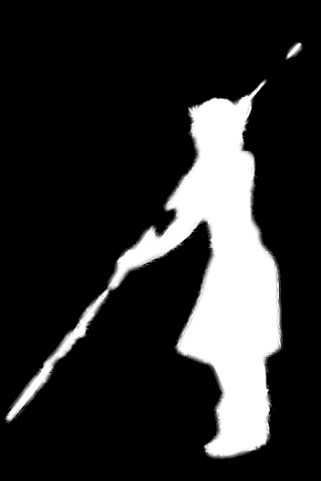} &  
    \includegraphics[height=.13\linewidth, width=.193\linewidth]{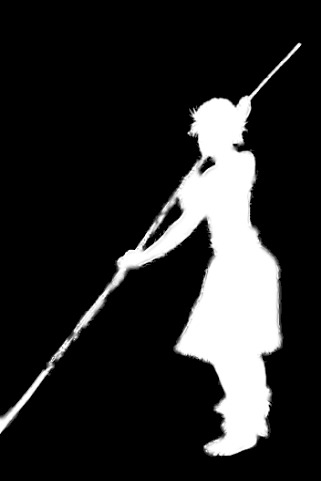} & 
    \includegraphics[height=.13\linewidth, width=.193\linewidth]{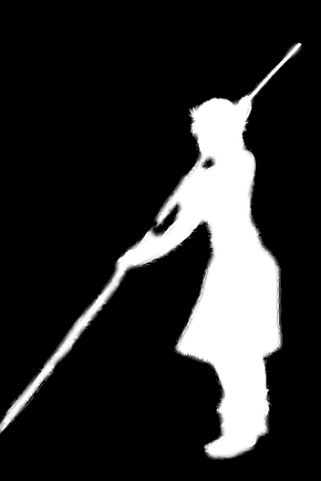} \\
    \end{tabular}
    \caption{ \textbf{Example segmentations at 20 clicks, of models trained with SBD, trained with the synthetic dataset alone, and trained with the synthetic set and finetuned on SBD}: Images are from the Berkeley dataset.}
    \label{fig:synthetic:comparisonsHigh}
\end{figure*}

The comparisons of the mean IoU over the GrabCut, Berkeley and SBD datasets for
these three scenarios is presented in
Figures~\ref{fig:SynthAblationIOUgraphs:GrabCut},~\ref{fig:SynthAblationIOUgraphs:Berkeley}
and ~\ref{fig:SynthAblationIOUgraphs:SBD} and examples of the segmentation results at 3 clicks and 20 clicks for the three training approaches are shown in Figure~\ref{fig:synthetic:comparisonsLow} and~\ref{fig:synthetic:comparisonsHigh}.
\begin{table}[t]\scriptsize
\begin{tabular}{lccc|ccc}
\hline
Method                              & \multicolumn{3}{c|}{GrabCut}                                                                                                                                                      & \multicolumn{3}{c}{Berkeley}                                                                                                        \\
                                    & \begin{tabular}[c]{@{}c@{}}\textit{NoC}\\  \textit{@ 90\%}\end{tabular} & \textit{AuC}   & \begin{tabular}[c]{@{}c@{}}\textit{\textgreater 99\%}\\ \textit{@ 20 clicks}\end{tabular} & \begin{tabular}[c]{@{}c@{}}NoC \\ @ 90\%\end{tabular} & AuC   & \begin{tabular}[c]{@{}c@{}}\textgreater 95\%\\ @ 20 clicks\end{tabular} \\
\oursItertrainEncoderFeedbackGF     & 2.54                                                                    & 0.963          & 62\%                                                                                   & 3.53                                                  & 0.942 & 81\%                                                                 \\
\oursItertrainEncoderFeedbackGFPT   & 1.96                                                                    & 0.968          & \textbf{74\%}                                                                          & 3.46                                                  & 0.943 & 82\%                                                                 \\
\oursItertrainEncoderFeedbackGFPTFT & \textbf{1.8}                                                            & \textbf{0.974} & 71\%                                                                                   & \textbf{3.04}                                         & \textbf{0.949} & \textbf{83\%}                                                        \\ \hline
\end{tabular}
\caption{Comparison of training with SBD alone(\oursItertrainEncoderFeedbackGF), with the synthetic dataset alone(\oursItertrainEncoderFeedbackGFPT), and with finetuning the synthetic model on the SBD set(\oursItertrainEncoderFeedbackGFPTFT). We measure the proportion of images that surpass 95\% \& 99\% accuracy at 20 clicks.}
\label{tab:synthetic:comparison}

\end{table}

What transpires from these results is that training on the synthetic dataset
helps considerably extracting fine details in the masks(see Table~\ref{tab:synthetic:comparison}). This is visible on the
examples of Figure~\ref{fig:synthetic:comparisonsLow} and
Figure~\ref{fig:synthetic:comparisonsHigh}. This can also be deduced from the
IoU plots on Berkeley and GrabCut, as both synthetic trained (1) and fine tuned
(2) versions outperform the SBD trained network (3) as the number of clicks
increases.

On the other hand, training on the SBD seems to help with the first few
clicks. This makes sense as SBD is designed for rough semantic shapes, whereas
our synthetic dataset is not specifically tailored for semantic classes.

We note that the difference between both training sets is most evident on the
SBD benchmark, where using the synthetic dataset actually negatively impacts the
performance. This makes sense as SBD has only 20 classes, thus training on the
SBD training set gives a decisive advantage.

\section{Discussion}

We have shown that our interactive segmentation method can reach levels of accuracy that can be as high as 99\% mIoU. What can be done to achieve higher levels of accuracy? One limitation is that binary segmentation is not well defined for natural images. Pixels at the object boundaries are never either foreground or background but always a mix of both  (\eg hair, fur, motion blur, defocus, \etc). This is exactly what natural matting is trying to solve by defining transparency masks. Thus, instead of interactive segmentation, we should probably aim for interactive matting instead, at least on high-end applications.

One other limitation we observed in this paper is that we are partly limited by the quality of existing training sets. The semantic segmentation datasets available for training and testing are based on real images but are of low quality. On the other hand, the synthetic datasets used in matting, and also in this paper, are of high quality but are not based on real images. There is thus room for much better training sets. The same problem is true for the test sets used in the three popular benchmarks. Going beyond 95\% accuracy on the SBD benchmark requires to match a ground truth that does not necessarily follow the actual object boundaries.

In the end, there seems to be two problems in one: rough segmentation and extracting fine details. It is difficult to solve for both in a single unified approach. We can point out to the BRS (\citet{jang2019interactive}), or FCTSFN (\citet{Hu2019ASegmentation}), which split their architectures into two parts: a core network that produces a rough segmentation and, appended to it, a refinement network to upsample the masks. Both networks are trained in stages. In our approach, we show that a single network can be used, but we still resort to a two-stage training when using the synthetic dataset. How to best schedule the training for rough and fine details seems to be key.

\begin{figure*}[p]
  \begin{tabular}[t]{lm{10em}}
     \resizebox{0.7\linewidth}{!}{\input{fig/SyntheticDataset/_synthetic_ablation_grabcut.pgf}} & 
                      {\vspace{-5em} \caption{Training with our Synthetic dataset. Mean IoU scores after $n$ clicks
                          for the {\bf GrabCut} testset~\cite{grabcutDB}.\label{fig:SynthAblationIOUgraphs:GrabCut}}}
                      \\
     \resizebox{0.7\linewidth}{!}{\input{fig/SyntheticDataset/_synthetic_ablation_berkeley.pgf}} &
                      {\vspace{-5em} \caption{Training with our Synthetic
                          dataset. Comparison of mean IoU scores after $n$ clicks
                          for the {\bf Berkeley} testset~\cite{BerkeleyDB}.\label{fig:SynthAblationIOUgraphs:Berkeley}}}
                      \\
     \resizebox{0.7\linewidth}{!}{\input{fig/SyntheticDataset/_synthetic_ablation_sbd.pgf}} &
                      {\vspace{-5em} \caption{Training with our Synthetic
                          dataset. Comparison of mean IoU scores after $n$ clicks
                          for the {\bf SBD} testset~\cite{BharathICCV2011}.\label{fig:SynthAblationIOUgraphs:SBD}}}
  \end{tabular}
\end{figure*}

\section{Conclusion}\label{sec:conclusion}

Existing deep-learning based interactive segmentation methods are of limited use
in professional photo editing applications as these methods plateau in accuracy
around 95\% and yet artists need to achieve more than 99\% accuracy. One reason
for this plateau is that they insufficiently leverage user interactions. We
proposed a novel single network architecture that better embeds the user
interactions into two separate streams. Using a click by click training regime
helps us improve the correlation between the click placements and the prediction
errors.  Our experiments show how each contribution improves the response to
local corrections and mIoU accuracy across the full range of clicks. In
comparison to existing approaches, our method achieves higher accuracy for all
20 measured clicks, across three benchmarks and we even achieve 99\% accuracy for 62\% of images on the GrabCut dataset within 20 clicks.

We also make the observation that the low quality of existing training sets limits the potential performance of current interactive networks. We show that introducing a more accurate synthetic training set can further improve the overall accuracy of our system, this model reaches 99\% accuracy for 74\% of images on the GrabCut dataset within 20 clicks.

\bibliographystyle{elsarticle-num-names}
\bibliography{longstrings,references}

\end{document}